\documentclass[journal]{IEEEtran}
\usepackage{xspace,amsmath,amssymb,amsthm,amsfonts,epsfig,syntonly,dsfont}
\usepackage{cite,bm,color,url,textcomp,array}
\usepackage{algorithmicx,algorithm,algpseudocode}
\usepackage{epstopdf,mathrsfs}
\usepackage{empheq,float,mdframed}
\usepackage{graphicx,balance,subcaption}

\newtheorem{lemma}{Lemma}
\newtheorem{corollary}{Corollary}
\newtheorem{theorem}{Theorem}

\theoremstyle{definition}\newtheorem{remark}{Remark}
\newmdtheoremenv{them}{Theorem}

\DeclareMathOperator*{\argmin}{arg\,min}

\def\baselinestretch{0.97}
\long\def\symbolfootnote[#1]#2{\begingroup
\def\thefootnote{\fnsymbol{footnote}}
\footnote[#1]{#2}\endgroup}
\psfull

\allowdisplaybreaks[2]

\begin{document}

\title{Bandit Convex Optimization for\\ Scalable and Dynamic IoT Management}
\author{
Tianyi Chen and Georgios B. Giannakis
\thanks {Work in this paper was supported by NSF 1509040, 1508993, and 1711471.}

\thanks{T. Chen and G. B. Giannakis are with the Department of Electrical and Computer Engineering and the Digital Technology Center, University of Minnesota, Minneapolis, MN 55455 USA. Emails: \{chen3827, georgios\}@umn.edu
}
}
\maketitle


\begin{abstract}

The present paper deals with online convex optimization involving both time-varying loss functions,  and time-varying constraints. 
The loss functions are not fully accessible to the learner, and instead only the function values (a.k.a. bandit feedback) are revealed at queried points.  
The constraints are revealed after making decisions, and can be instantaneously violated, yet they must be satisfied in the long term. This setting fits nicely the emerging online network tasks such as fog computing in the Internet-of-Things (IoT), where online decisions must flexibly adapt to the changing user preferences (loss functions), and the temporally unpredictable availability of resources (constraints). 
Tailored for such human-in-the-loop systems where the loss functions are hard to model, 
a family of bandit online saddle-point (BanSaP) schemes are developed, which adaptively adjust the online operations based on (possibly multiple) bandit feedback of the loss functions, and the changing environment. 
Performance here is assessed by: i) \emph{dynamic regret} that generalizes the widely used static regret; and, ii) \emph{fit} that captures the accumulated amount of constraint violations. 
Specifically, BanSaP is proved to simultaneously yield sub-linear dynamic regret and fit, provided that the best dynamic solutions vary slowly over time. Numerical tests in fog computation offloading tasks corroborate that our proposed BanSaP approach offers competitive performance relative to existing approaches that are based on gradient feedback.

\end{abstract}


\section{Introduction}
Internet-of-Things (IoT) envisions an intelligent infrastructure of networked smart devices offering task-specific monitoring and control services \cite{samie2016b}. 
Leveraging advances in embedded systems, contemporary IoT devices are featured with \emph{small-size} and \emph{low-power} designs, but their computation and communication capabilities are limited. 
A prevalent solution during the past decade was to move computing, control, and storage resources to the remote cloud (a.k.a. data centers). 
Yet, the cloud-based IoT architecture is challenged by high latency due to directly communications with the cloud, which certainly prevents real-time applications \cite{chiang2016}.
Along with other features of IoT, such as \emph{extreme heterogeneity} and \emph{unpredictable dynamics}, the need arises for innovations in network design and management to allow for adaptive online service provisioning, subject to stringent delay constraints \cite{lee2017}. 

From the network design vantage point, \emph{fog} is viewed as a promising architecture for IoT that distributes computation, communication, and storage closer to the end IoT users, along the cloud-to-things continuum \cite{chiang2016}. 
In the fog computing paradigm, service provisioning starts at the network edge, e.g., smartphones, and high-tech routers, and only a portion of tasks will be offloaded to the powerful cloud for further processing (a.k.a. computation offloading) \cite{samie2016,mach2017,wang2017}.
Existing approaches for computation offloading either focus on time-invariant static settings, or, rely on stochastic optimization approaches such as Lyapunov optimization to deal with time-varying cases; see \cite{mao2017} and references therein. 
Nevertheless, static settings cannot capture the changing IoT environment, and the stationarity commonly assumed in stochastic optimization literature may not hold in practice, especially when the stochastic process involves human participation as in IoT.
From the management perspective, online network control, which is robust to non-stationary dynamics and amenable to \emph{light-weight} implementations, remains a largely uncharted territory \cite{mach2017,mao2017}.

Indeed, the \emph{primary goal} of this paper is an algorithmic pursuit of online network optimization suitable for emerging tasks in IoT. 
Focusing on such algorithmic challenges, online convex optimization (OCO) is a promising methodology for
sequential tasks with well-documented merits, especially when the sequence of convex costs varies in an unknown and possibly adversarial manner \cite{zinkevich2003}.
Aiming to empower traditional fog management policies with OCO, most available OCO works benchmark algorithms with a \textit{static regret}, which measures the difference of costs (a.k.a. losses) between the online solution and the best static solution in hindsight \cite{hazan2007,duchi2011jmlr}.
However, static regret is not a comprehensive performance metric in dynamic settings such as those encountered with IoT \cite{jadbabaie2015}.

Recent works extend the
analysis of static to that of \textit{dynamic regret} \cite{hall2015,jadbabaie2015}, but they deal with time-invariant constraints that cannot be violated instantaneously. 
Tailored for fog computing setups that need flexible adaptation of online decisions to dynamic resource availability, OCO with time-varying constraints was first studied in \cite{chen2017tsp},  along with its adaptive variant in \cite{chen2017iot}, and the optimal regret bound in this setting was first established in \cite{neely2017}.  
Yet, the approaches in \cite{chen2017iot,chen2017tsp,neely2017} remain operational under the premise that the loss functions are \emph{explicitly known}, or, their gradients are readily available. 
Clearly, none of these two assumptions can be easily satisfied in IoT settings, because i) the loss function capturing user dissatisfaction, e.g., service latency or reliability, is hard to model in dynamic environments; and, ii) even if modeling is possible in theory, the low-power IoT devices may not afford the complexity of running statistical learning tools such as deep neural networks ``on-the-fly.''   
 
In this context, targeting a gradient-free light-weight solution, alternative online schemes have been advocated leveraging point-wise values of loss functions (partial-information feedback) rather than their gradients (full-information feedback). They are termed bandit convex optimization (BCO) in machine learning \cite{flaxman2005,agarwal2010,shamir2017,bubeck2012}, or referred as zeroth-order schemes in optimization circles \cite{duchi2015,nesterov2017}.  
While \cite{flaxman2005,agarwal2010,duchi2015,shamir2017} and \cite{nesterov2017} employed on BCO with time-invariant constraints that cannot be violated instantaneously, the \textit{long-term} effect of such instantaneous violations was studied in \cite{mahdavi2012}, where the focus is still on static regret and time-invariant constraints.
Building on full-information precursors \cite{chen2017tsp,chen2017iot,neely2017}, the present paper broadens the scope of BCO to the regime with \emph{time-varying constraints}, and proposes a class of online algorithms termed online bandit saddle-point (BanSaP) approaches. 
With an eye on managing IoT with limited information, our contribution is the incorporation of long-term and time-varying constraints to expand the scope of BCO, as well as an improved regret-fit tradeoff relative to that in \cite{mahdavi2012}; see a summary in Table \ref{tab:comp}.

\begin{table}\addtolength{\tabcolsep}{1pt}
\centering \caption{A summary of related works on OCO/BCO}\label{tab:comp} \vspace{-0cm}
    \begin{tabular}{ |c *{3}{|c}|}
    \hline
Reference   & Benchmark & 	Constraints  & Feedback \\ \hline
\cite{zinkevich2003,hazan2007,duchi2011jmlr}    & Static          & Fixed and strict          & Gradient                   \\
\cite{hall2015,jadbabaie2015}         & Dynamic         & Fixed and strict   & Gradient                          \\
\cite{neely2017} & Static &Varying and long-term & Gradient \\
\cite{chen2017tsp,chen2017iot} & Dynamic &Varying and long-term & Gradient \\
\cite{mahdavi2012} & Static &Fixed and long-term & Grad./Fun. value \\
\cite{flaxman2005,agarwal2010,duchi2015,nesterov2017,shamir2017,bubeck2012}    & Static        & Fixed and strict     & Function value                    \\ \hline
 This work & Dynamic        &  Varying and long-term              & Function value                \\
 \hline
    \end{tabular} \vspace{-0.4cm}
\end{table}

In a nutshell, relative to existing works, the main contributions of the
present paper are summarized as follows.

\textbf{c1)} We generalize the standard BCO framework with only
time-varying costs \cite{flaxman2005,agarwal2010}, to account for
both time-varying costs and constraints. Performance here is established relative to the best dynamic
benchmark, via metrics that we term dynamic regret and fit (Section III).

\textbf{c2)} We develop a class of BanSaP algorithms to tackle this novel BCO
problem, and analytically establish that BanSaP solvers yield
simultaneously optimal sub-linear dynamic regret and fit, given that
the accumulated variations of per-slot minimizers are known to grow sub-linearly with time (Section IV).

\textbf{c3)} Our BanSaP algorithms are applied to computation offloading tasks emerging in IoT management, and simulations demonstrate that the BanSaP solvers have comparable performance relative to full-information alternatives (Section V).


\emph{Notation}. $(\cdot)^{\top}$ stands for
vector and matrix transposition, and $\|\mathbf{x}\|$ denotes the $\ell_2$-norm of a vector $\mathbf{x}$. Inequalities for vectors $\mathbf{x} > \mathbf{0}$, and the projection $[\mathbf{a}]^+:=\max\{\mathbf{a},\mathbf{0}\}$ are entry-wise.

\section{Bandit Online Learning with Constraints}\label{sec.LTOCO}

In this section, a generic BCO formulation with long-term and time-varying constraints will be introduced, along with its real-world application in IoT management. 

\subsection{Online learning with constraints under partial feedback}

Before introducing BCO with long-term constraints, we begin with the classical BCO setting, where constraints are
time-invariant, and must be strictly satisfied \cite{flaxman2005,agarwal2010,bubeck2012}. 
Akin to its full-information counterpart \cite{zinkevich2003,hazan2007}, BCO can be viewed as a repeated game between a learner and nature. 
Consider that time is discrete and indexed by $t$. Per slot $t$, a learner
selects an action $\mathbf{x}_t$ from a convex set
${\cal X}\subseteq\mathbb{R}^d$, and subsequently nature chooses a loss function $f_t(\cdot):
\mathbb{R}^d\rightarrow \mathbb{R}$ through which the learner incurs a
loss $f_t(\mathbf{x}_t)$. 
The convex feasible set ${\cal X}$ is a-priori known and fixed over the entire time horizon. 
Different from the OCO setup, at the end of each slot, only the value of $f_t(\mathbf{x}_t)$ rather than the form of $f_t(\mathbf{x})$ is revealed to the learner in BCO.  
Although this standard BCO setting is appealing to various applications
such as online end-to-end routing \cite{awerbuch2004} and task assignment 
\cite{kao2016}, it does not account for potential variations of (possibly unknown) constraints, and does not deal with constraints that can possibly be satisfied in the long
term rather than a slot-by-slot basis \cite{mahdavi2012,chen2017tsp,neely2017}.

Online optimization with time-varying and long-term constraints is
well motivated for applications from power control in wireless communication \cite{neely2010}, geographical load balancing in cloud networks \cite{chen2016tsp,chen2017tsp}, to computation offloading in fog computing \cite{sardellitti2015,chen2016ton}. 
Motivated by these dynamic network management tasks, our recent works \cite{chen2017tsp,chen2017iot} studied OCO with time-varying constraints in \emph{full information} setting, where the gradient feedback is available. 
Complementing \cite{chen2017tsp} and \cite{chen2017iot}, the present paper broadens the applicability of BCO to the regime with time-varying long-term constraints.

Specifically, we consider that per slot $t$, a learner selects an action $\mathbf{x}_t$ from a known and fixed convex set ${\cal X}\subseteq\mathbb{R}^d$, and then nature chooses not only a loss
function $f_t(\cdot): \mathbb{R}^d\rightarrow \mathbb{R}$, but also
a time-varying penalty function
$\mathbf{g}_t(\cdot): \mathbb{R}^d\rightarrow \mathbb{R}^N$. The later gives rise to the time-varying constraint
$\mathbf{g}_t(\mathbf{x})\leq \mathbf{0}$, which is driven by the
unknown application-specific dynamics. 
Similar to the standard BCO setting, only the value of $f_t(\mathbf{x}_t)$ at the queried point $\mathbf{x}_t$ is revealed to the learner here; but different from the standard BCO setting, besides  ${\cal X}$, the constraint
$\mathbf{g}_t(\mathbf{x})\leq \mathbf{0}$ needs to be carefully taken care of. 
And the fact that $\mathbf{g}_t$ is unknown to the learner when performing her/his decision, makes it impossible to satisfy in every time slot. 
Hence, a more realistic goal here
is to find a sequence of solutions $\{\mathbf{x}_t\}$ that minimizes the aggregate loss, and ensures that the
constraints $\{\mathbf{g}_t(\mathbf{x}_t)\leq \mathbf{0}\}$ are
satisfied in the long term on average. 
Specifically, extending the BCO framework \cite{flaxman2005,agarwal2010,shamir2017} to
accommodate such time-varying constraints, we consider the following online optimization problem
\begin{empheq}[box=\fbox]{align}\label{eq.prob}
\min_{\{\mathbf{x}_t\in {\cal X},\forall t\}} ~\sum_{t=1}^T f_t(\mathbf{x}_t)~~~\text{s. to}~\sum_{t=1}^T \mathbf{g}_t(\mathbf{x}_t) \leq \mathbf{0}
\end{empheq}
where $T$ is the entire time horizon, $\mathbf{x}_t\in\mathbb{R}^d$ is
the decision variable, $f_t$ represents the cost function,
$\mathbf{g}_t:=[g_t^1,\ldots,g_t^N]^{\top}$ denotes the constraint
function with $n$th entry $g_t^n(\cdot):\mathbb{R}^d\rightarrow
\mathbb{R}$, and ${\cal X}\in \mathbb{R}^d$ is a convex set. 
In the current setting, we assume that only the values of loss function are available at queried points since e.g., its complete form related to user experience is hard to approximate, but the constraint function is revealed to the learner as it represents measurable physical requirements e.g., power budget, and data flow conservation constraints.     
Before the algorithm development in Section \ref{sec.BanSaP} and performance analysis in Section \ref{sec.perform}, we will introduce a motivating example of fog computing in IoT.

\subsection{Motivating setup: mobile fog computing in IoT}\label{subsec.fog}
\begin{figure}[t]
\centering
\vspace{-0cm}
\includegraphics[width=0.49\textwidth]{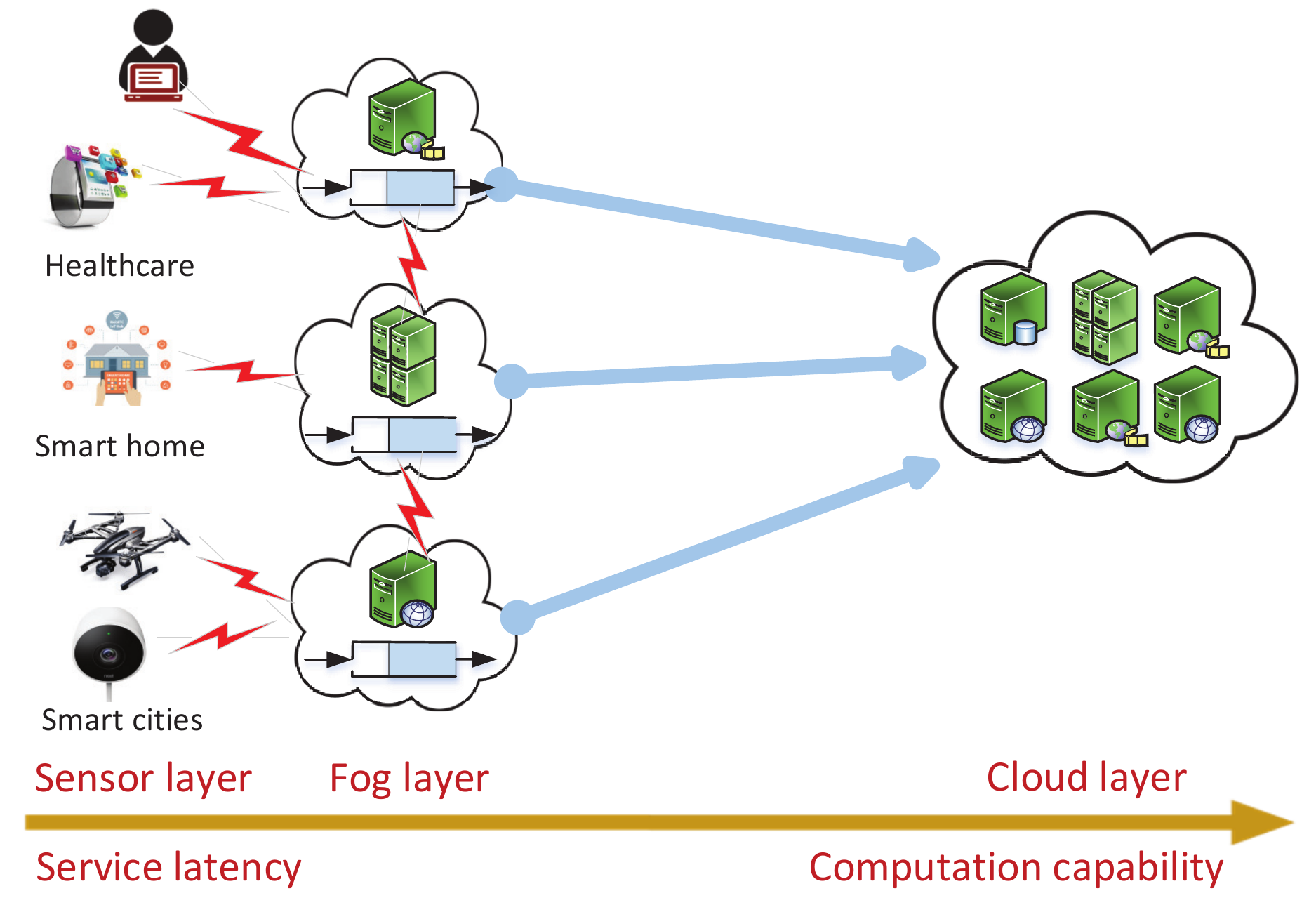}
\vspace{-0.5cm} \caption{A diagram of hierarchical fog computing framework.} \label{fig:system}
\vspace{-0.5cm}
\end{figure}

The online computational offloading task of fog computing in IoT \cite{mao2017,mach2017,samie2016} takes the form of BCO with long-term constraints \eqref{eq.prob}. 
Consider a mobile network with a sensor layer, a fog layer, and a cloud layer \cite{lee2017,chiang2016}.
The sensor layer contains heterogeneous low-power IoT devices (e.g., wearable watches and smart cameras), which do not have enough computational capability, and usually offload their collected data to the local fog nodes (e.g., smartphones and high-tech routers) in the fog layer for further processing \cite{huang2017}. 
The fog layer consists of $N$ nodes in the set ${\cal N}:=\{1,\ldots,N\}$ with moderate processing capability; thus, part of workloads will be collaboratively processed by the local fog servers to meet the stringent latency requirement, and the rest will be offloaded to the remote data center in the cloud layer \cite{mach2017}; also see Fig. \ref{fig:system}. 

Per time $t$, each fog node $n$ collects data requests $b_t^n$ from all its nearby sensors. 
Once receiving these requests, node $n$ has \textit{three options}: i) offloading the amount $z_t^n$ to the remote data center; ii) offloading the amount $y_t^{nk}$ to each of its nearby node $k$ for collaborative computing; and, iii) locally processing the amount $y_t^{nn}$ according to its resource availability. 
The optimization variable $\mathbf{x}_t$ in this case consists of the cloud offloading, local offloading, and local processing amounts; i.e., $\mathbf{x}_t\!:=\![z_t^1,\ldots,z_t^N,y_t^{11},\ldots,y_t^{1N},\ldots,y_t^{N1},\ldots,y_t^{NN}]^{\top}$.
Assuming that each fog node has a data queue to buffer unserved workloads, the instantaneously served workloads (offloading plus processing) is not necessarily equal to the data arrival rate.
Instead, a long-term constraint is common to ensure that the cumulative amount of served workloads is no less than the arrived amount at each node $n$ over time \cite{neely2010} 
\begin{equation}\label{eq.long-contrs}
	\sum_{t=1}^T g_t^n(\mathbf{x}_t):=\sum_{t=1}^T\Bigg(b_t^n\!+\!\sum_{k\in{\cal N}_n^{\rm in}}\! y_t^{kn}\!-\!\!\sum_{k\in{\cal N}_n^{\rm out}}\! y_t^{nk}-z_t^n-y_t^{nn}\Bigg) \!\leq 0
\end{equation}
where ${\cal N}_n^{\rm in}$ and ${\cal N}_n^{\rm out}$ represent the sets of fog nodes with in-coming links to node $n$ and those with out-going links from node $n$, respectively.
The bandwidth limit of communication link (e.g., wireline) from fog node $n$ to the remote cloud is
$\bar{z}^n$; the limit of the transmission link (e.g., wireless) from node $n$ to its neighbor $k$ is
$\bar{y}^{nk}$, and the computation capability of node $n$ is
$\bar{y}^{nn}$. 
With $\bar{\mathbf{x}}$ collecting all the aforementioned limits, the feasible region can be expressed by $\mathbf{x}_t \!\in\! {\cal X}\!:=\!\{\mathbf{0}\!\leq\! \mathbf{x}_t\!\leq\!
\bar{\mathbf{x}}\}$.

Performance is assessed by the user dissatisfaction of the online processing and offloading decisions, e.g., aggregate delay \cite{samie2016b,lee2017}.  
Specifically, as the computation delay is usually negligible for data centers with thousands of high-performance servers, the latency for cloud offloading amount $z_t^n$ is mainly due to the communication delay, which is denoted as a time-varying cost $c_t^n(z_t^n)$ depending on the unpredictable network congestion during slot $t$. 
Likewise, the communication delay of the local offloading decision $y_t^{nk}$ from node $n$ to a nearby node $k$ is denoted as $c_t^{nk}(y_t^{nk})$, but its magnitude is much lower than that of cloud offloading.
Regarding the processing amount $y_t^{nn}$, its latency comes from the computation delay due to its limited computational capability, which is presented as a time-varying function $h_t^n(y_t^{nn})$ capturing the dynamic CPU capability during the computing processes.
Per slot $t$, the network delay 
$f_t(\mathbf{x}_t)$ aggregates the computation delay at all nodes plus the communication delay at all links, namely
\begin{equation}\label{eq.netcost}
\!f_t(\mathbf{x}_t)\!:=\!\!\sum_{n\in
{\cal N}}\!\!\Bigg(\!\!\underbrace{c_t^n(z_t^n)+\textstyle\sum_{k\in {\cal N}_n^{\rm out}}\!c_t^{nk}(y_t^{nk})}_{\rm communication}+\!\!\underbrace{h_t^{n}
(y_t^{nn})}_{\rm computation}\!\Bigg).\!\!\!\!
\end{equation}
Clearly, the explicit form of functions $c_t^n(\cdot)$, $c_t^{nk}(\cdot)$, and $h_t^n(\cdot)$ is \emph{unknown} to the network operator due to the unpredictable traffic patterns \cite{awerbuch2004}; but they are convex (thus $f_t(\mathbf{x}_t)$ is convex) with respect to their arguments, which implies that the marginal computation/communication latency is increasing as the offloading/processing amount grows.    

Aiming to minimize the accumulated network delay while serving all the IoT workloads in the long term, the optimal offloading strategy in this mobile network is the solution of the following online optimization problem (cf. \eqref{eq.netcost})
\begin{align}\label{eq.apply}
\min_{\{\mathbf{x}_t\in {\cal X},\forall t\}} \, &\sum_{t=1}^T f_t(\mathbf{x}_t),~
\text{s. to}~~\eqref{eq.long-contrs}~{\rm for~}n=1,\ldots,N.
\end{align}
Comparing to the generic form \eqref{eq.prob}, we consider an online fog computing problem in \eqref{eq.apply}, where the loss (network latency) function $f_t(\cdot)$ and the data requests $\{b_t^n\}$ within slot $t$ are not known when making the offloading and local processing decision $\mathbf{x}_t$; after performing $\mathbf{x}_t$, only the value of $f_t(\mathbf{x}_t)$ (a.k.a. loss) as well as the measurements $\{b_t^n\}$ are revealed to the network operator. 
In this example, measuring $\{b_t^n\}$ is tantamount to knowing the constraint function $g_t^n(\cdot)$ in \eqref{eq.long-contrs}. 
Therefore, \eqref{eq.apply} is in the form of \eqref{eq.prob}.

\section{Online Bandit Saddle-Point Methods}\label{sec.BanSaP}

To solve the problem in Section \ref{sec.LTOCO}, an online saddle-point method is revisited first, before developing its bandit variants for network optimization with only partial feedback.

\subsection{Online saddle-point approach with gradient feedback}
Several works have studied the OCO setup with time-varying long-term constraints (cf. \eqref{eq.prob}), including \cite{chen2017tsp,neely2017}, and the recent variant \cite{chen2017iot} incorporating with adaptive stepsizes. 
Consider now the per-slot problem \eqref{eq.prob}, which
contains the current objective $f_t(\mathbf{x})$, the current
constraint $\mathbf{g}_t(\mathbf{x})\leq \mathbf{0}$, and a
time-invariant feasible set ${\cal X}$. 
With $\bm{\lambda}\in
\mathbb{R}^N_+$ denoting the Lagrange multiplier associated with
the time-varying constraint, the online Lagrangian of
\eqref{eq.prob} can be expressed as
\begin{align}\label{eq.Lam}
{\cal L}_t(\mathbf{x},\bm{\lambda}):=f_t(\mathbf{x})+\bm{\lambda}^{\top}\mathbf{g}_t(\mathbf{x}).
\end{align}

Serving as a basis for developing the bandit approaches, we next revisit the online saddle-point scheme with full-information \cite{neely2017}, that is also equivalent to \cite{chen2017tsp} when $\mathbf{g}_t(\mathbf{x})$ is linear.
Specifically, given the primal iterate $\mathbf{x}_t$ and the dual iterate
$\bm{\lambda}_t$ at each slot $t$, the next decision
$\mathbf{x}_{t+1}$ is generated by
\begin{equation}\label{eq.primal-min}
       \mathbf{x}_{t+1}\!\in \arg\min_{\mathbf{x}\in{\cal X}} \nabla_{\mathbf{x}}^{\top}{\cal L}_t(\mathbf{x}_t,\bm{\lambda}_t)(\mathbf{x}-\mathbf{x}_t)+\frac{1}{2\alpha}\left\|\mathbf{x}-\mathbf{x}_t\right\|^2
\end{equation}
where $\alpha$ is a pre-defined constant, and $\nabla_{\mathbf{x}}
{\cal L}_t(\mathbf{x}_t,\bm{\lambda}_t)=\nabla f_t(\mathbf{x}_t)+\nabla^{\top} \mathbf{g}_t(\mathbf{x}_t)\bm{\lambda}_t$ is the gradient of ${\cal L}_t(\mathbf{x},\bm{\lambda}_t)$ with respect to (w.r.t.) the primal variable $\mathbf{x}$ at $\mathbf{x}=\mathbf{x}_t$. 
The minimization \eqref{eq.primal-min} admits the closed-form solution, given by
\begin{equation}\label{eq.primal}
       \mathbf{x}_{t+1}={\cal P}_{\cal X}\!\left(\mathbf{x}_t-\alpha\nabla_{\mathbf{x}}{\cal L}_t(\mathbf{x}_t,\bm{\lambda}_t)\right)
\end{equation}
where ${\cal P}_{\cal X}(\mathbf{y})\!:=\!\argmin_{\mathbf{x}\in{\cal X}}\|\mathbf{x}-\mathbf{y}\|^2$ denotes the projection operator.
In addition, the dual update takes the modified online gradient ascent form
\begin{equation}\label{eq.dual}
        \bm{\lambda}_{t+1}=\left[\bm{\lambda}_t+\mu (\mathbf{g}_t(\mathbf{x}_t)+\nabla^{\top} \mathbf{g}_t(\mathbf{x}_t)(\mathbf{x}_{t+1}-\mathbf{x}_t))\right]^{+}
\end{equation}
where $\mu$ is a positive stepsize, and
$\nabla_{\bm{\lambda}}{\cal L}_t(\mathbf{x}_t,\bm{\lambda}_t)=\mathbf{g}_t(\mathbf{x}_t)$ is
the gradient of ${\cal L}_t(\mathbf{x}_t,\bm{\lambda})$ w.r.t. $\bm{\lambda}$ at $\bm{\lambda}=\bm{\lambda}_t$. 
Note that \eqref{eq.dual} is a modified gradient update since the dual variable is updated along the first-order approximation of $\mathbf{g}_t(\mathbf{x}_{t+1})$ at the previous iterate $\mathbf{x}_t$ rather than $\mathbf{g}_t(\mathbf{x}_t)$ used in \cite{chen2017tsp}, which will be critical in our subsequent analytical derivations. 

To perform the online saddle-point recursion \eqref{eq.primal}-\eqref{eq.dual} however, the gradient $\nabla f_t(\mathbf{x})$ and the constraint $\mathbf{g}_t(\mathbf{x})$ should be known to the learner at each slot $t$. 
When the gradient of $f_t(\mathbf{x})$ (or its explicit form) is unknown as it is in our setup, additional effort is needed. 
In this context, the systematic design of the online \emph{bandit} saddle-point (BanSaP) methods will be leveraged to extend the online saddle-point method to the regime where gradient information is unavailable or computationally costly. 

\begin{figure}[t]
\hspace{-0.2cm}
\includegraphics[width=0.5\textwidth]{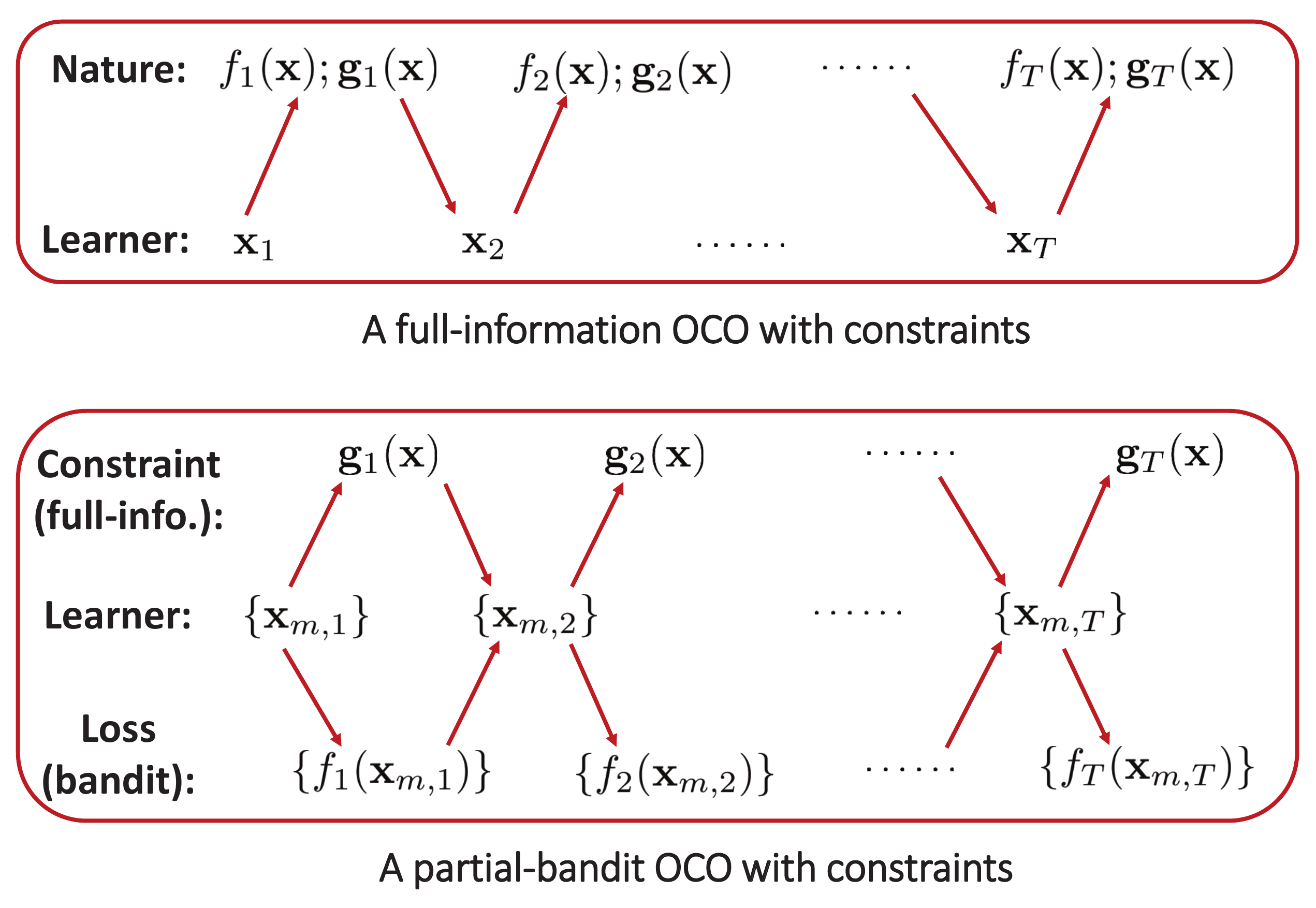}
\vspace{-0.4cm} \caption{A comparison of OCO with full/partial-bandit feedback.} \label{fig:ocodiag}
\vspace{-0.4cm}
\end{figure}

\subsection{BanSaP with one-point partial feedback}\label{subsec.BanSaP-1}
The key idea behind BCO is to construct (possibly stochastic) gradient estimates using the limited \emph{function value} information \cite{flaxman2005,agarwal2010,duchi2015,nesterov2017,shamir2017}. 
Depending on system variability, the online learner can afford one or multiple loss function evaluations (partial-information feedback) per time slot  \cite{nesterov2017,agarwal2010,duchi2015}. 
Intuitively, the performance of a bandit algorithm will improve if multiple evaluations are available per time slot; see Fig. \ref{fig:ocodiag} for a comparison of full- versus partial-information feedback settings. 

To begin with, we consider the case where the learner can only observe the function value of $f_t(\mathbf{x})$ at a single point per slot $t$. The crux here is to construct a (possibly unbiased) estimate of the gradient using this single piece of feedback.
Interestingly though, a stochastic gradient estimate of $f_t(\mathbf{x})$ can be obtained by one point \emph{random} function evaluation \cite{flaxman2005}. 
The intuition can be readily revealed from the one-dimensional case ($d=1$): For a binary random variable $u$ taking values $\{-1,1\}$ equiprobable, and a small constant $\delta>0$, the idea of forward differentiation implies that the derivative $f_t'$ at $x$ can be approximated by 
\begin{equation}
f_t'(x)\approx \frac{f_t(x+\delta )-f_t(x-\delta)}{2\delta}=\mathbb{E}_{u}\left[\frac{u}{\delta}f_t(x+\delta u)\right]
\end{equation}
where the approximation is due to $\delta>0$, and the equality follows from the definition of expectation. Hence, $f_t(x+\delta u)u/\delta$ can serve as a stochastic estimator of $f_t'(x)$ based only single function evaluation $f_t(x+\delta u)$.
Generalizing this approximation to high dimensions, with a random vector $\mathbf{u}$ drawn from the unit sphere (a.k.a. the surface of a unit ball), the scaled function evaluation at a perturbed point $\mathbf{x}+\delta\mathbf{u}$ yields an estimate of the gradient $\nabla f_t(\mathbf{x})$, given by \cite{flaxman2005}
\begin{equation}\label{eq.grad1}
\nabla f_t(\mathbf{x})\approx \mathbb{E}_{\mathbf{u}}\left[\frac{d}{\delta}f_t(\mathbf{x}+\delta \mathbf{u})\mathbf{u}\right]:=\mathbb{E}_{\mathbf{u}}\left[\hat{\nabla}^1 f_t(\mathbf{x})\right]
\end{equation}
where we define one-point gradient $\hat{\nabla}^1 f_t(\mathbf{x}):=\frac{d}{\delta}f_t(\mathbf{x}+\delta \mathbf{u})\mathbf{u}$.

Building upon this intuition, consider a bandit version of the online saddle-point iteration, for which the primal update becomes (cf. \eqref{eq.primal})
\begin{equation}\label{eq.primal2}
       \hat{\mathbf{x}}_{t+1}={\cal P}_{(1-\gamma){\cal X}}\!\left(\hat{\mathbf{x}}_t-\alpha\hat{\nabla}_{\mathbf{x}}^1{\cal L}_t(\hat{\mathbf{x}}_t,\bm{\lambda}_t)\right)
\end{equation}
where $(1-\gamma){\cal X}:=\{(1-\gamma)\mathbf{x}:\mathbf{x}\in{\cal X}\}$ is a subset of ${\cal X}$,  $\gamma\in[0,1)$ is a pre-selected constant depending on $\delta$, and the one-point Langragian gradient is given by (cf. \eqref{eq.grad1})
\begin{equation}\label{eq.grad2}
\hat{\nabla}_{\mathbf{x}}^1{\cal L}_t(\hat{\mathbf{x}}_t,\bm{\lambda}_t):=\hat{\nabla}^1 f_t(\hat{\mathbf{x}}_t)+\nabla^{\top} \mathbf{g}_t(\hat{\mathbf{x}}_t)\bm{\lambda}_t.\!
\end{equation}
In the full-information case, $\mathbf{x}_t$ in \eqref{eq.primal} is the learner's action, but in the bandit case the learner's action is $\mathbf{x}_{1,t}:=\hat{\mathbf{x}}_t+\delta \mathbf{u}_t$, which is the point for function evaluation but not $\hat{\mathbf{x}}_t$ in \eqref{eq.primal2}. 
Furthermore, the projection is performed on a smaller convex set $(1-\gamma){\cal X}$ in \eqref{eq.primal2}, which ensures feasibility of the perturbed $\mathbf{x}_{1,t}\in {\cal X}$. 
Similar to the full-information case \eqref{eq.dual}, the dual update of BanSaP is given by
\begin{equation}\label{eq.dual2}
        \bm{\lambda}_{t+1}=\left[\bm{\lambda}_t+\mu (\mathbf{g}_t(\hat{\mathbf{x}}_t)+\nabla^{\top} \mathbf{g}_t(\hat{\mathbf{x}}_t)(\hat{\mathbf{x}}_{t+1}-\hat{\mathbf{x}}_t))\right]^{+}
\end{equation}
where $\mu$ is again the stepsize, and the learning iterate $\hat{\mathbf{x}}_t$ rather than the actual decision $\mathbf{x}_t$ is used in this update. 
Compared with the gradient-based recursions \eqref{eq.primal}-\eqref{eq.dual}, the updates \eqref{eq.primal2}-\eqref{eq.dual2} with one-point bandit feedback do not increase computation or memory requirements, and thus provide a light-weight surrogate for gradient-free online bandit network optimization.

\begin{algorithm}[t]
\caption{BanSaP for OCO with time-varying constraints}\label{algo1}
\begin{algorithmic}[1]
\State \textbf{Initialize:} primal iterate $\hat{\mathbf{x}}_1$, dual
iterate $\bm{\lambda}_1$, parameters $\delta$ and $\gamma$, and proper stepsizes $\alpha$ and $\mu$.
\For {$t=1,2\dots$} 
\State The learner plays the perturbed actions $\{\mathbf{x}_{m,t}\}_{m=1}^M$ {\color{white} ccc\,}based on the learning iterate $\hat{\mathbf{x}}_t$.
\State The nature reveals the losses $\{f_t(\mathbf{x}_{m,t})\}_{m=1}^M$ at queried {\color{white} ccc\,}points, and the constraint function $\mathbf{g}_t(\mathbf{x})$.
\State The learner updates the primal variable $\hat{\mathbf{x}}_{t+1}$ by \eqref{eq.primal2} {\color{white} ccc\,}with the gradient estimated by \eqref{eq.grad2} for $M=1$, or, \eqref{eq.grad3} {\color{white} ccc\,}for $M=2$, otherwise, by \eqref{eq.gradM}.
\State The learner updates the dual variable $\bm{\lambda}_{t+1}$ via \eqref{eq.dual2}.
\EndFor
\end{algorithmic}
\end{algorithm}
\setlength{\textfloatsep}{8pt}

\subsection{BanSaP with multipoint partial feedback}\label{subsec.BanSaP-m}
Featuring a simple update given minimal information, the BanSaP with one-point bandit feedback is suitable for fast-varying environments, where multiple function evaluations are impossible.  
As shown later in Sections \ref{sec.perform} and \ref{sec.num}, the theoretical and empirical performance of BanSaP with single-point evaluation is degraded relative to the full-information case. 

To improve the performance of BanSaP with one-point feedback, we will first rely on two-point function evaluation at each slot \cite{duchi2015}, and then generalize to multipoint evaluation. 
Intuitively, this approach is justified when the underlying dynamics are slow, e.g., when the load and price profiles in power grids are piece-wise stationary. 
In this case, each slot can be further divided into multiple mini-slots, and one query is performed per mini-slot, over which the loss function and the constraints do not change. 
Compared to \eqref{eq.primal2}-\eqref{eq.dual2}, the key difference is that the one-point estimate in \eqref{eq.grad2} is replaced by
\begin{align}\label{eq.stograd3}
	\hat{\nabla}^2 f_t(\hat{\mathbf{x}}_t):=\frac{d}{2\delta}\big(f_t(\hat{\mathbf{x}}_t+\delta \mathbf{u}_t)-f_t(\hat{\mathbf{x}}_t-\delta \mathbf{u}_t)\big)\mathbf{u}_t
\end{align}
where the function values are evaluated on two points around the learning iterate $\hat{\mathbf{x}}_t$, namely, $\mathbf{x}_{1,t}:=\hat{\mathbf{x}}_t+\delta \mathbf{u}_t$ and $\mathbf{x}_{2,t}:=\hat{\mathbf{x}}_t-\delta \mathbf{u}_t$ with $\mathbf{u}_t$ again drawn uniformly from the unit sphere $\mathbb{S}:=\{\mathbf{u}\in \mathbb{R}^d:\|\mathbf{u}\|=1\}$. The primal update becomes $\hat{\mathbf{x}}_{t+1}={\cal P}_{(1-\gamma){\cal X}}\big(\hat{\mathbf{x}}_t-\alpha\hat{\nabla}_{\mathbf{x}}^2{\cal L}_t(\hat{\mathbf{x}}_t,\bm{\lambda}_t)\big)$,  
with Lagrangian gradient 
\begin{align}\label{eq.grad3}
\hat{\nabla}_{\mathbf{x}}^{2}{\cal L}_t(\hat{\mathbf{x}}_t,\bm{\lambda}_t):=\hat{\nabla}^2 f_t(\hat{\mathbf{x}}_t)	+\nabla \mathbf{g}_t(\hat{\mathbf{x}}_t)^{\!\top}\bm{\lambda}_t.
\end{align}

Similar to the one-point case, it is instructive to consider the two-point gradient estimate in the one-dimensional case ($d=1$), where the expectation of the differentiation term in \eqref{eq.stograd3} approximates well the derivative of $f_t$ at $\hat{x}_t$; that is, 
\begin{align}
	&\mathbb{E}_{u}\left[\frac{u_t}{2\delta}\left(f_t(\hat{x}_t+\delta u_t)-f_t(\hat{x}_t-\delta u_t)\right)\right]\nonumber\\
	=& \frac{1}{2\delta}\left(f_t(\hat{x}_t+\delta )-f_t(\hat{x}_t-\delta)\right)\approx f_t'(\hat{x}_t)
\end{align}
where the equality follows because the random variable $u_t$ takes values $\{-1,1\}$ equiprobable. 

Relative to the one-point feedback case, the advantage of the two-point feedback is variance reduction in the gradient estimator. 
Specifically, the second moment of the stochastic gradient can be uniformly bounded, $\mathbb{E}[\|\frac{d}{2\delta}\big(f_t(\hat{\mathbf{x}}_t+\delta \mathbf{u}_t)-f_t(\hat{\mathbf{x}}_t-\delta \mathbf{u}_t)\big)\mathbf{u}_t\|^2]\leq d^2G^2$, where $G$ is the Lipschitz constant of $f_t(\mathbf{x})$. This is in contrast to the one-point feedback where the second moment is inversely proportional to $\delta$, since $\mathbb{E}[\frac{d}{\delta}\|f_t(\hat{\mathbf{x}}_t+\delta \mathbf{u}_t)\mathbf{u}_t\|^2]\leq d^2 F^2/{\delta^2}$, with $F$ denoting an upper-bound of $f_t(\mathbf{x})$. The proof of this argument can be found in the Appendix (Lemma \ref{app-lemma0}).
In fact, a bias-variance tradeoff emerges in the one-point case, but not in the two-point case. 
This subtle yet critical difference will be responsible for an improved performance of BanSaP with two-point feedback, and its stable empirical performance, as will be seen later.

With the insights gained so far, the next step is to endow the BanSaP with more than two function evaluations \cite{agarwal2010}. 
With $M>2$ points, the gradient estimator is obtained by querying the function values over $M$ points in the neighborhood of $\hat{\mathbf{x}}_t$. These points include $\mathbf{x}_{m,t}\!\!:=\!\hat{\mathbf{x}}_t\!+\delta \mathbf{u}_{m,t},\,1\!\leq\! m \!\leq\! M-1$, and the learning iterate $\mathbf{x}_{m,t}:=\hat{\mathbf{x}}_t$, where $\mathbf{u}_{m,t}$ is independently drawn from $\mathbb{S}$. Specifically, the gradient becomes (cf. \eqref{eq.primal2})
\begin{align}\label{eq.gradM}
\!\!\!&\hat{\nabla}_{\mathbf{x}}^{M}{\cal L}_t(\hat{\mathbf{x}}_t,\bm{\lambda}_t):=\\
\!\!\!&\frac{d}{\delta (M-1)}\!\!\sum_{m=1}^{M-1}\!\!\big(f_t(\hat{\mathbf{x}}_t\!+\!\delta \mathbf{u}_{m,t})\!-\!f_t(\hat{\mathbf{x}}_t)\big)\mathbf{u}_{m,t}+\nabla \mathbf{g}_t(\hat{\mathbf{x}}_t)^{\!\top}\bm{\lambda}_t\nonumber
\end{align}
where we define the $M$-point stochastic gradient as $\hat{\nabla}^{M}\! f_t(\hat{\mathbf{x}}_t)\!:=\!\frac{d}{\delta (M-1)}\!\sum_{m=1}^{M-1}\!\big(f_t(\hat{\mathbf{x}}_t\!+\!\delta \mathbf{u}_{m,t})\!-\!f_t(\hat{\mathbf{x}}_t)\big)\mathbf{u}_{m,t}$. 
At the price of extra computations, simulations will validate that the BanSaP with multipoint feedback enjoys improved performance. 
The family of the BanSaP approaches with one- or multiple-point feedback is summarized in Algorithm \ref{algo1}. 

\begin{remark}[Sampling schemes]
The BanSaP solvers here adopt uniform sampling for gradient estimation, meaning $\mathbf{u}$ is drawn uniformly from the unit sphere. However, other sampling rules can be incorporated without affecting the order of regret bounds derived later.
For example, one can sample $\mathbf{u}$ from the canonical basis of a $d$-dimensional space uniformly at random \cite{agarwal2010}, or, sample $\mathbf{u}$ from a normal distribution \cite{nesterov2017}. 
The effectiveness of these schemes will be tested using simulations.  
\end{remark}

\section{Performance analysis}\label{sec.perform}
In this section, we will introduce pertinent metrics to evaluate BanSaP algorithms in the online bandit learning with long-term constraints, and rigorously analyze the performance of the proposed algorithms.

\subsection{Optimality and feasibility metrics}
With regard to performance of BCO schemes, static regret is a common metric, under time-invariant and strictly satisfied constraints, which measures the difference
between the aggregate loss and that of the best
fixed solution in hindsight
\cite{agarwal2010,flaxman2005}.
Extending the definition of static regret to accommodate $M$-point function evaluations and time-varying constraints, let us first consider 
\begin{align}\label{eq.static-reg}
    {\rm Reg}^{\rm s}_T:=\frac{1}{M}\sum_{t=1}^T \sum_{m=1}^M \mathbb{E}\left[f_t(\mathbf{x}_{m,t})\right]-\sum_{t=1}^T f_t(\mathbf{x}^*)
\end{align}
where the actual loss per slot is averaged over the losses of $M$ actions (queried points), $\mathbb{E}$ is taken over the sequence of random actions (due to $\delta \mathbf{u}$ perturbations), and the best static solution is $\mathbf{x}^*\in\arg\min_{\mathbf{x}\in {\cal X}} \,\sum_{t=1}^T f_t(\mathbf{x});~\text{s. to}~\mathbf{g}_t(\mathbf{x}) \leq \mathbf{0},\;\forall t$.
A BCO algorithm yielding a
sub-linear regret implies that the algorithm
is ``on average'' no-regret \cite{mahdavi2012}; or, in other words, asymptotically not worse than the
best fixed solution $\mathbf{x}^*$. 
Though widely used, the \textit{static regret} relies on a rather coarse benchmark, which is not as useful in dynamic IoT settings.
Specifically, the gap between the loss of the best static and that of the best dynamic
benchmark is as large as ${\cal O}(T)$ \cite{besbes2015}.

In response to the quest for improved benchmarks in this dynamic setup with constraints, two metrics
are considered here: \textit{dynamic regret} and \textit{dynamic fit}. The notion of dynamic regret has been recently adopted in
\cite{jadbabaie2015,hall2015} to assess performance of online algorithms under
time-invariant constraints. For our BCO setting of
\eqref{eq.prob}, we adopt
\begin{align}\label{eq.dyn-reg}
    {\rm Reg}^{\rm d}_T:=\frac{1}{M}\sum_{t=1}^T \sum_{m=1}^M \mathbb{E}\left[f_t(\mathbf{x}_{m,t})\right]-\sum_{t=1}^T f_t(\mathbf{x}_t^*)
\end{align}
where $\mathbb{E}$ is again taken over the sequence of random actions, and the benchmark is now formed via a sequence of best dynamic
solutions $\{\mathbf{x}_t^*\}$ for the instantaneous cost
minimization problem subject to the instantaneous constraint,
namely
\begin{equation}\label{eq.realtime-prob}
    \mathbf{x}_t^*\in\arg\min_{\mathbf{x}\in {\cal X}} \; f_t(\mathbf{x})~~~\text{s. to}~~~\mathbf{g}_t(\mathbf{x}) \leq \mathbf{0}.
\end{equation}
Quantitatively, the dynamic regret is always larger
than the static regret, i.e., ${\rm
Reg}^{\rm s}_T \leq {\rm Reg}^{\rm d}_T$, since 
$\sum_{t=1}^Tf_t(\mathbf{x}^*)$ is always no smaller than
$\sum_{t=1}^Tf_t(\mathbf{x}^*_t)$ according to the definitions of
$\mathbf{x}^*$ and $\mathbf{x}^*_t$.
Hence, a sub-linear dynamic regret implies a sub-linear static regret, but not vice versa.

Regarding feasibility of decisions generated by a BCO algorithm, the notion of
\textit{dynamic fit} will be used to measure the
accumulated violation of constraints \cite{mahdavi2012}, that is
\begin{align}\label{eq.dyn-fit}
    {\rm Fit}^{\rm d}_T:=\Bigg\|\Bigg[\frac{1}{M}\sum_{t=1}^T \sum_{m=1}^M \mathbf{g}_t(\mathbf{x}_{m,t})\Bigg]^+\Bigg\|.
\end{align}
Note that the dynamic fit is zero if the accumulated
violation $\frac{1}{M}\sum_{t=1}^T \sum_{m=1}^M \mathbf{g}_t(\mathbf{x}_{m,t})$ is entry-wise
less than zero. Hence, enforcing
$\frac{1}{M}\sum_{t=1}^T \sum_{m=1}^M \mathbf{g}_t(\mathbf{x}_{m,t}) \!\leq\! \mathbf{0}$ is
different from restricting $\mathbf{x}_t$ to meet
$\frac{1}{M} \sum_{m=1}^M \mathbf{g}_t(\mathbf{x}_{m,t})\leq \mathbf{0}$ in every slot.
While the latter readily implies the former, the long-term constraint implicitly assumes that the instantaneous constraint violations can be compensated by the later strictly feasible decisions, and thus allows adaptation of online decisions to the unknown dynamics.

Under this broader BCO setup, an ideal online algorithm is the
one that achieves both sub-linear dynamic regret and sub-linear
dynamic fit. A sub-linear dynamic regret implies ``no-regret''
relative to the clairvoyant dynamic solution on the long-term
average; i.e., $\lim_{T\rightarrow \infty}{{\rm Reg}_{T}^{\rm
d}}/{T}=0$; and a sub-linear dynamic fit indicates that the online
strategy is also feasible on average; i.e., $\lim_{T\rightarrow
\infty}{{\rm Fit}_{T}^{\rm d}}/{T}=0$. Unfortunately, the sub-linear
dynamic regret is not achievable in general, even when the time-varying constraint in \eqref{eq.prob}
is absent \cite{besbes2015}.
Therefore, we aim at designing an online strategy that generates a sequence $\{\mathbf{x}_{m,t}\}$ ensuring sub-linear dynamic regret and fit, under the suitable conditions on the underlying dynamics.

\vspace{-0.3cm}

\subsection{Main results}
Before formally analyzing the dynamic regret and fit for BanSaP, we assume that the following conditions are satisfied.

\noindent\textbf{(as1)} \emph{For every $t$, the functions 
$f_t(\mathbf{x})$ and 
$\mathbf{g}_t(\mathbf{x})$ are convex.}

\noindent\textbf{(as2)} \emph{Function $f_t(\mathbf{x})$ is bounded over the set ${\cal X}$, meaning $|f_t(\mathbf{x})|\leq F,\, \forall \mathbf{x}\in {\cal X}$; while $f_t(\mathbf{x})$ and $g_t^n(\mathbf{x})$ have bounded gradients; that is, $\|\nabla
f_t(\mathbf{x})\|\leq G$, and $\max_n\|\nabla g_t^n(\mathbf{x})\|\leq G$.}

\noindent\textbf{(as3)} \emph{For a small constant $\gamma$, there exists a constant $\eta>0$, and an interior point $\tilde{\mathbf{x}}\in (1-\gamma){\cal X}$ such that $\mathbf{g}_t(\tilde{\mathbf{x}})\leq -\eta\mathbf{1},\;\forall t$.}

\noindent\textbf{(as4)} \emph{With $\mathbb{B}\!:=\!\{\mathbf{x}\in\mathbb{R}^d:\|\mathbf{x}\|\leq 1\}$ denoting the unit ball, there exist constants $0<r\leq R$ such that $r\mathbb{B}\subseteq{\cal X}\subseteq R\mathbb{B}$.}

Assumptions (as1)-(as2) are typical in OCO with both full- and partial-information feedback
\cite{hazan2007,mahdavi2012,flaxman2005}; (as3) is Slater's condition modified for our BCO setting, which guarantees the existence of a bounded Lagrange multiplier \cite{bertsekas1999} in the constrained optimization; and, (as4) requires the action set to be bounded within a ball that contains the origin. 
When (as4) appears to be restrictive, it is tantamount to assuming ${\cal X}$ is compact and has a nonempty interior, because one can always apply an affine transformation (a.k.a. reshaping) on ${\cal X}$ to satisfy (as4); see \cite[Section 3.2]{flaxman2005}. 

Under these assumptions, we are on track to first provide upper bounds for the dynamic regret, and the dynamic fit of the BanSaP solver with one-point feedback. 
\begin{theorem}[one-point feedback]\label{Them1}
Suppose that (as1)-(as4) are satisfied, and consider the parameters $\alpha$, $\mu$, $\delta$, $\gamma$ defined in \eqref{eq.primal2}-\eqref{eq.dual2}, and constants $F$, $G$, $r$, $R$ defined in (as2)-(as4). 
If the dual variable is initialized by $\bm{\lambda}_1=\mathbf{0}$, then the BanSaP with one-point feedback in \eqref{eq.primal}-\eqref{eq.dual} has dynamic regret bounded by
\begin{align}\label{Them.dyn-reg}
  {\rm Reg}^{\rm d}_T\leq \frac{R}{\alpha}V(\mathbf{x}_{1:T}^*)&+\!\frac{R^2}{2\alpha}\!+\!\frac{d^2G^2R^2\alpha T}{\delta^2}+\!2G\delta T\nonumber\\
  &+\gamma GRT\left(1+\|\bar{\bm{\lambda}}\|\right)\!+\!2\mu G^2R^2T
\end{align}
where $\|\bar{\bm{\lambda}}\|:=\max_t \|\bm{\lambda}_t\|$, and the accumulated variation of the per-slot minimizers $\mathbf{x}^*_t$ in \eqref{eq.realtime-prob} is given by
\begin{equation}\label{eq.var-min}
 V(\mathbf{x}_{1:T}^*):=\sum_{t=1}^T \|\mathbf{x}^*_t-\mathbf{x}^*_{t-1}\|.	
\end{equation}
In addition, the dynamic fit defined in \eqref{eq.dyn-fit} is bounded by
\begin{align}\label{Them.fit}
  \!\!\!\!\!   {\rm Fit}^{\rm d}_T\!\leq\! \frac{\|\bar{\bm{\lambda}}\|}{\mu}\!+\!\frac{G^2\sqrt{N}T}{2\beta}&+\!\delta G\sqrt{N}T\!\nonumber\\
        &+\!\beta \sqrt{N}T\!\left(\!\frac{\alpha^2 d^2F^2 }{\delta^2}\!+\!\alpha^2 G^2\|\bar{\bm{\lambda}}\|^2\!\right)\!\!
\end{align}
where $\beta>0$ is a pre-selected constant. 
Furthermore, if we choose the stepsizes as $\alpha=\mu={\cal O}(T^{-\frac{3}{4}})$, and the parameters $\delta={\cal O}(T^{-\frac{1}{4}})$, $\beta=T^{\frac{1}{4}}$ and $\gamma=\delta/r$, then the online decisions generated by BanSaP are feasible, i.e., $\mathbf{x}_{1,t}\in{\cal X}$; and also yield the following dynamic regret and fit
\begin{empheq}[box=\fbox]{align}\label{Them.dyn-reg1}
  {\rm Reg}^{\rm d}_T\!=\! {\cal O}\Big(V(\mathbf{x}_{1:T}^*)T^{\frac{3}{4}}\Big){\rm ~~and~~}{\rm Fit}^{\rm d}_T 
	={\cal O}\big(T^{\frac{3}{4}}\big).
\end{empheq}
\end{theorem}	

\begin{IEEEproof}
See Appendix \ref{app-prf1}.	
\end{IEEEproof}

For BanSaP with one-point feedback, Theorem \ref{Them1} asserts that its dynamic regret and fit are upper-bounded by some
constants depending on the those parameters, the time horizon, and the accumulated variation of per-slot minimizers.
Interestingly, the crucial constant $\delta$ controlling the perturbation of random actions appears in both the denominator and numerator of \eqref{Them.dyn-reg} and \eqref{Them.fit}, which correspond to the variance and bias of the gradient estimator. 
Therefore, simply setting a small $\delta$ will not only reduce the bias, but it will also boost the variance - a clear manifestation of the that is known as bias-variance tradeoff in BCO \cite{shamir2017}.
Optimally choosing parameters implies that the dynamic fit is sub-linearly growing, and the dynamic regret is sub-linear given that the variation of the per-slot minimizer is slow enough; i.e., $V(\mathbf{x}_{1:T}^*)=\mathbf{o}(T^{\frac{1}{4}})$. 

Regarding BanSaP with two-point feedback, we can prove the following result that parallels Theorem \ref{Them1}.
\begin{theorem}[two-point feedback]\label{Them2}
Consider the assumptions and the definitions of constants in Theorem \ref{Them1}. 
If the dual variable is initialized by $\bm{\lambda}_1=\mathbf{0}$, then BanSaP with two-point feedback has dynamic regret bounded by
\begin{align}\label{Them2.dyn-reg}
  {\rm Reg}^{\rm d}_T\leq \frac{R}{\alpha}V(\mathbf{x}_{1:T}^*)+\!\frac{R^2}{2\alpha}&+\!2\mu G^2R^2T\!+\!\alpha d^2G^2 T\nonumber\\
  &+\!\gamma GRT(1+\|\bar{\bm{\lambda}}\|)\!+\!2\delta GT
\end{align}
and has dynamic fit in \eqref{eq.dyn-fit} bounded by
\begin{align}\label{Them2.fit}
  \!\!\!\!{\rm Fit}^{\rm d}_T\!\leq\! \frac{\|\bar{\bm{\lambda}}\|}{\mu}\!+\!\frac{G^2\sqrt{N}T}{2\beta}\!&+\!\delta G\sqrt{N}T\nonumber\\
  &+\!\beta \sqrt{N}T\!\left(\alpha^2 d^2G^2\!+\!\alpha^2 G^2\|\bar{\bm{\lambda}}\|^2\right)\!.\!
\end{align}
In this case, if we choose the stepsizes as $\alpha=\mu={\cal O}(T^{-\frac{1}{2}})$, and set the parameters as $\beta=T^{\frac{1}{2}}$,  $\delta={\cal O}(T^{-1})$, and $\gamma=\delta/r$, then the online decisions generated by BanSaP are feasible, and its dynamic regret and fit are bounded by
\begin{empheq}[box=\fbox]{align}\label{Them2.dyn-reg1}
 	  {\rm Reg}^{\rm d}_T\!=\! {\cal O}\Big(V(\mathbf{x}_{1:T}^*)T^{\frac{1}{2}}\Big){\rm~~and~~}{\rm Fit}^{\rm d}_T 
	={\cal O}\big(T^{\frac{1}{2}}\big)
\end{empheq}
where $V(\mathbf{x}_{1:T}^*)$ is the accumulated
variation of the per-slot minimizers $\mathbf{x}^*_t$ in \eqref{eq.var-min}.
\end{theorem}
\begin{IEEEproof}
See Appendix \ref{app-prf2}.	
\end{IEEEproof}

Comparing with the bounds in \eqref{Them.dyn-reg} and \eqref{Them.fit}, the perturbation constant $\delta$ only appears in the numerator of \eqref{Them2.dyn-reg} and \eqref{Them2.fit} because our gradient estimator here replies on two points. 
In this case, the additional function evaluation allows BanSaP to choose an arbitrarily small $\delta$ to minimize the bias of stochastic gradient, without increasing its variance. This observation is aligned with those in BCO without long-term constraints \cite{agarwal2010,shamir2017}. 
Furthermore, Theorem \ref{Them2} establishes that the dynamic regret and fit are sub-linear if $V(\mathbf{x}_{1:T}^*)=\mathbf{o}(T^{\frac{1}{2}})$, which markedly improves those in Theorem \ref{Them1} under one-point feedback. 

For the case of BanSaP with $M>2$ points, slightly improved bounds can be proved without changing the order of regret and fit, but they are omitted here for brevity. 
In addition, the bounds in Theorems \ref{Them1} and \ref{Them2} can be achieved without any knowledge of $V(\mathbf{x}_{1:T}^*)$.
When the order of $V(\mathbf{x}_{1:T}^*)$ is known, or, can be estimated a-priori, tighter regret and fit bounds can be obtained by adjusting stepsizes accordingly.
Formally, we can arrive at the following corollary.
\begin{corollary}\label{ref.coro0}
	Under the conditions of Theorems \ref{Them1} and \ref{Them2}, suppose that there exists a constant $\rho\in[0,1)$ such that the variation satisfies $V(\mathbf{x}_{1:T}^*)=\mathbf{o}(T^{\rho})$. If the stepsizes of BanSaP with one-point feedback are chosen as $ \alpha=\mu={\cal O}\big(T^{\frac{3}{4}(\rho-1)}\big)$, 
    and the parameters are $\delta={\cal O}(T^{\frac{1}{4}(\rho-1)})$, $\beta=T^{\frac{3}{4}(1-\rho)}$,  and $\gamma=\delta/r$, 
	 then the dynamic regret and fit in \eqref{Them.dyn-reg1} become
	\begin{equation}\label{opt-reg}
     {\rm Reg}^{\rm d}_T\!=\!{\cal O}\!\left(T^{\frac{1}{4}(\rho+3)}\right){\rm~~and~~}{\rm Fit}^{\rm d}_T={\cal O}\left(T^{\frac{1}{4}(\rho+3)}\right).
    \end{equation}
Likewise, if the stepsizes of BanSaP with two-point feedback are chosen such that $\alpha=\mu={\cal O}\big(T^{\frac{1}{2}(\rho-1)}\big)$, 
    and the parameters are $\delta={\cal O}(T^{\frac{1}{2}(\rho-1)})$, $\beta=T^{\frac{1}{2}(1-\rho)}$,  and $\gamma=\delta/r$, 
	 then the dynamic regret and fit in \eqref{Them.dyn-reg1} become
	\begin{equation}\label{opt-reg}
     {\rm Reg}^{\rm d}_T\!=\!{\cal O}\!\left(T^{\frac{1}{2}(\rho+1)}\right){\rm~~and~~}{\rm Fit}^{\rm d}_T={\cal O}\left(T^{\frac{1}{2}(\rho+1)}\right).
    \end{equation}
\end{corollary}

Apparently, Corollary \ref{ref.coro0} implies that sub-linear dynamic regret and fit are both possible, provided that the accumulated variation of the minimizers is growing sub-linearly ($\rho<1$), and it is available to the learner in advance. It provides valuable insights for choosing optimal stepsizes in dynamic environments.
Specifically, adjusting stepsizes to match the variability of the environment is the key to achieving the optimal dynamic regret and fit. Intuitively, when the variation is fast (large $\rho$), slowly decaying stepsizes (thus larger stepsizes) can better track the potential changes; and vice versa.

\begin{remark}[Optimal regret]
As a special case of Theorems \ref{Them1} and \ref{Them2}, by confining $\mathbf{x}_1^*=\cdots=\mathbf{x}_T^*$ so that $V(\mathbf{x}_{1:T}^*)=0$, the dynamic regret bounds \eqref{Them.dyn-reg1} and \eqref{Them2.dyn-reg1} reduce to the static ones, which correspond to ${\cal O}(T^{\frac{3}{4}})$ in the one-point feedback case, and to ${\cal O}(\sqrt{T})$ in the two-point case. 
This pair of bounds markedly improves the \emph{regret versus fit tradeoff} in \cite{mahdavi2012}, and matches the order of regret in \cite{flaxman2005}, and \cite{agarwal2010,duchi2015}, which are the best possible ones that can be achieved by \emph{efficient} algorithms even in the BCO setup without the long-term constraints.
\end{remark}

\begin{remark}[Dynamic regret]
Theorems \ref{Them1}, \ref{Them2} and Corollary \ref{ref.coro0} extend the dynamic regret analysis in \cite{hall2015,jadbabaie2015,chen2017tsp} to the regime of \emph{bandit} online learning with long-term \emph{time-varying} constraints.
Interestingly though, in the BCO setting of our interest, sub-linear dynamic regret and fit are possible to achieve when the per-slot minimizer \textit{does not vary on average}, that is, $V(\mathbf{x}_{1:T}^*)$ is sub-linearly growing with $T$. 
\end{remark}

\begin{algorithm}[t]
\caption{BanSaP for fog computation offloading}\label{algo2}
\begin{algorithmic}[1]
\State \textbf{Initialize:} primal iterates $\{\hat{y}_1^{nk}\}$ and $\{\hat{z}_1^n\}$, dual
iterate $\bm{\lambda}_1$, parameters $\delta$ and $\gamma$, and proper stepsizes $\alpha$ and $\mu$.
\For {$t=1,2\dots$} 
\For {$m=1,\dots,M$} 
\State Fog nodes perform \emph{perturbed} offloading decisions {\color{white} ccccccc}to cloud $\{z_{m,t}^n\}$, to neighbor edges $\{y_{m,t}^{nk}\}$, and {\color{white} ccccccc}locally process $\{y_{m,t}^{nn}\}$ based on $\hat{\mathbf{x}}_t$.
\EndFor
\State Fog nodes observe the (possibly multiple) losses to {\color{white} ccc\,}update \eqref{eq.dist-net} with stochastic gradients obtained via \eqref{eq.grad-net}. 
\State Fog nodes observe the actual user demands from IoT 
{\color{white} ccc\,}devices to update the dual variables \eqref{eq.dual-gd1}. 
\EndFor
\end{algorithmic}
\end{algorithm}
\setlength{\textfloatsep}{8pt}

\vspace{-0.3cm}

\section{Numerical Tests}\label{sec.num}

In this section, we demonstrate how the fog computation offloading task can benefit from our novel BanSaP solvers.

\subsection{BanSaP for fog computation offloading}\label{subsec.OMEC}
Recall that the computation offloading problem \eqref{eq.apply} is in the form of \eqref{eq.prob}. 
Therefore,
the BanSaP solver of Section \ref{sec.BanSaP} can be
customized to solve \eqref{eq.apply} in an \textit{online} fashion, with
provable performance and feasibility guarantees.

Specifically, with
$\mathbf{g}_t(\mathbf{x}_t)$ as in \eqref{eq.long-contrs} and $f_t(\mathbf{x}_t)$ as in \eqref{eq.netcost},
the primal update \eqref{eq.primal} boils down to a simple closed-form gradient update amenable to decentralized implementation; the cloud offloading amount at node $n$ is
\begin{subequations}\label{eq.dist-net}
\begin{equation}\label{eq.primal-gd1}
  \hat{z}_{t+1}^n=\left[\hat{z}_t^n-\alpha \big(\hat{\nabla} c^n_t(\hat{z}_t^n)-\lambda_t^n\big)\right]_{0}^{\bar{z}^n}
\end{equation}
and the offloading amount from node $n$ to node $k$ is given by
\begin{equation}\label{eq.primal-gd2}
\hat{y}_{t+1}^{nk}=\left[\hat{y}_t^{nk}\!-\!\alpha\big(\hat{\nabla} c^{nk}_t(\hat{y}_t^{nk})-\lambda_t^n+\lambda_t^k\big)\right]_{0}^{\bar{y}^{nk}}\!
\end{equation}
while the local processing decision at node $n$ is generated by
\begin{equation}\label{eq.primal-gd3}
 \hat{y}_{t+1}^{nn}=\left[\hat{y}_t^{nn}-\alpha\big(\hat{\nabla} h^n_t(\hat{y}_t^{nn})-\lambda_t^n\big)\right]_{0}^{\bar{y}^{nn}}
\end{equation}
\end{subequations}
where $\alpha$ is chosen according to Theorems \ref{Them1} and \ref{Them2}. 
Using two-point feedback $(M=2)$ as an example, the gradients involved in \eqref{eq.dist-net} can be estimated as
\begin{subequations}\label{eq.grad-net}
\begin{equation}\label{eq.grad-mec}
\hat{\nabla}^2 c^n_t(\hat{z}_t^n)\!:=\!\frac{d}{2\delta}\Big(f_t\big(\underbrace{\hat{\mathbf{x}}_t+\delta \mathbf{u}_t}_{\mathbf{x}_{1,t}})-f_t(\underbrace{\hat{\mathbf{x}}_t-\delta \mathbf{u}_t}_{\mathbf{x}_{2,t}}\big)\Big)u_t(\hat{z}^n)
\end{equation}
and with respect to the offloading variable, as 
\begin{equation}\label{eq.grad-mec2}
\hat{\nabla}^2 c^{nk}_t(\hat{y}_t^{nk})\!:=\!\frac{d}{2\delta}\big(f_t(\hat{\mathbf{x}}_t+\delta \mathbf{u}_t)-f_t(\hat{\mathbf{x}}_t-\delta \mathbf{u}_t)\big)u_t(\hat{y}^{nk})
\end{equation}
and with respect to the local processing variable, as
\begin{equation}\label{eq.grad-mec3}
\hat{\nabla}^2 h^n_t(\hat{y}_t^{nn})\!:=\!\frac{d}{2\delta}\big(f_t(\hat{\mathbf{x}}_t+\delta \mathbf{u}_t)-f_t(\hat{\mathbf{x}}_t-\delta \mathbf{u}_t)\big)u_t(\hat{y}^{nn})
\end{equation}
\end{subequations}
where $u_t(\hat{z}^n)$, $u_t(\hat{y}^{nk})$, and $u_t(\hat{y}^{nn})$ represent the corresponding entries of the random vector $\mathbf{u}_t\in\mathbb{R}^{|\cal E|}$ at slot $t$.

The dual update \eqref{eq.dual} at each node $n$ reduces to
 \begin{equation}\label{eq.dual-gd1}
\lambda_{t+1}^n\!=\!\!\Bigg[\lambda_t^n\!+\mu \Bigg(b_t^n+\!\!\sum_{k\in{\cal N}_n^{\rm in}} \!\!\hat{y}_{t+1}^{kn}-\!\!\!\sum_{k\in{\cal N}_n^{\rm out}}\!\! \hat{y}_{t+1}^{nk}\!-\hat{z}_{t+1}^n\!-\hat{y}_{t+1}^{nn}\Bigg)\!\Bigg]^{+}\!\!\!\!\!
\end{equation}
where $\mu$ is chosen according to Theorems \ref{Them1} and \ref{Them2}. 
Intuitively, to guarantee completion of the service requests, the dual variable increases (increasing penalty) when there is instantaneous service residual, and decreases when over-serving incurs in  the mobile-edge computing systems.
Following its generic form in Algorithm \ref{algo1}, BanSaP for online fog computation offloading tasks, is summarized in Algorithm \ref{algo2}.

%

\subsection{Numerical experiments}
Consider the fog computing task in Section \ref{subsec.fog} with $N=10$ nodes and a cloud center. 
Each fog node has an outgoing link to the cloud, and two outgoing links to two nearby fog nodes for local collaborative computing. 
For a communication link offloading loads from node $n$ to $k$, the offloading limit is $\bar{y}^{nk}\!=\!10$, the local computation limit at node $n$ is $\bar{y}^{nn}\!=\!50$, and the fog-cloud offloading limits $\{\bar{z}^n\}$ are all set to $100$.
The online cost (a.k.a. service latency) in \eqref{eq.netcost} is specified by
\begin{equation}
	f_t(\mathbf{x}_t)\!:=\!\sum_{n\in
{\cal N}}\!\Big(e^{p_t^n z_t^n}\!+\!\textstyle\sum_{k\in {\cal N}_n^{\rm out}}l^{nk}y_t^{nk}+l^{nn}(y_t^{nn})^2\Big)
\end{equation}
where $p_t^n\!=0.015\sin(\pi t/96)+0.05,\,n\!\in\!{\cal N}\backslash \{4,5\}$, $p_t^n\!=0.045\sin(\pi t/96)+0.15,\,n\!\in\!\{4,5\}$, and the local coefficients are set to $l^{nk}=8/\bar{y}^{nk}$ and $l^{nn}=8/\bar{y}^{nn}$. 
Regarding the data arrival rate $b_t^n$, it is generated according to $b_t^n\!=\!q^n\sin(\pi t/96)+\nu_t^n$, with $q^n$ and $\nu_t^n$ uniformly distributed over $[40,50]$ and $[45,55]$ for $n\!\in\!{\cal N}\backslash \{1,2,3\}\bigcup\{4,5\}$, and $q^n\!\in\![32,40],\nu_t^n\in[36,44],\,n\!\in\! \{1,2,3\}$, and $q^n\!\in\![20,25], \nu_t^n\!\in\![22.5,27.5],\,n\!\in\!\{4,5\}$.
Notice that the scales of $p_t^n$ and $b_t^n$ vary between nodes, mimicking heterogeneity of real IoT systems; and the periods of $p_t^n$ and $b_t^n$ correspond to a 24-hour interval with slot duration 7.5 minutes. 
When the parameters of BanSaP need to be slightly adjusted in each test, they are set to $\gamma=0.05$, and $\delta=4$ for with $M=1$, and $\delta=0.05$ for $M\geq 2$.

\begin{figure}[t]
\centering
\vspace{-0.4cm}
\hspace{-0.2cm}
\includegraphics[height=0.32\textwidth]{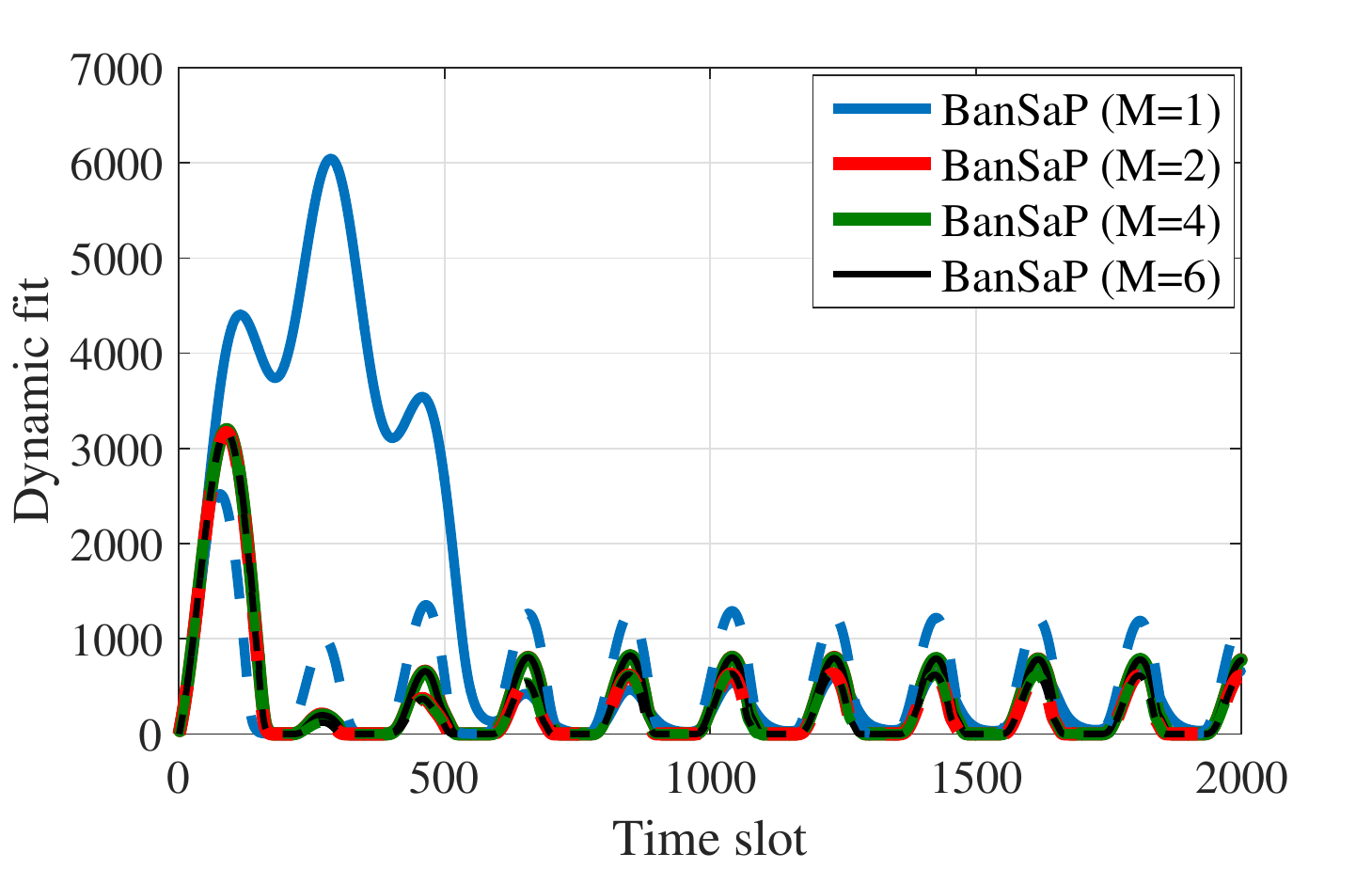}
\vspace{-0.2cm}
\caption{Effect of sampling schemes and number of feedback on dynamic fit. Solid lines: BanSaP with uniformly sampling from a unit sphere (uniform sampling). Dashed lines: BanSaP with randomly sampling from standard basis (coordinate sampling).}
\label{Fig.fit1}
\end{figure}

\begin{figure}[t]
\centering
\vspace{-0.4cm}
\hspace{-0.2cm}
\includegraphics[height=0.32\textwidth]{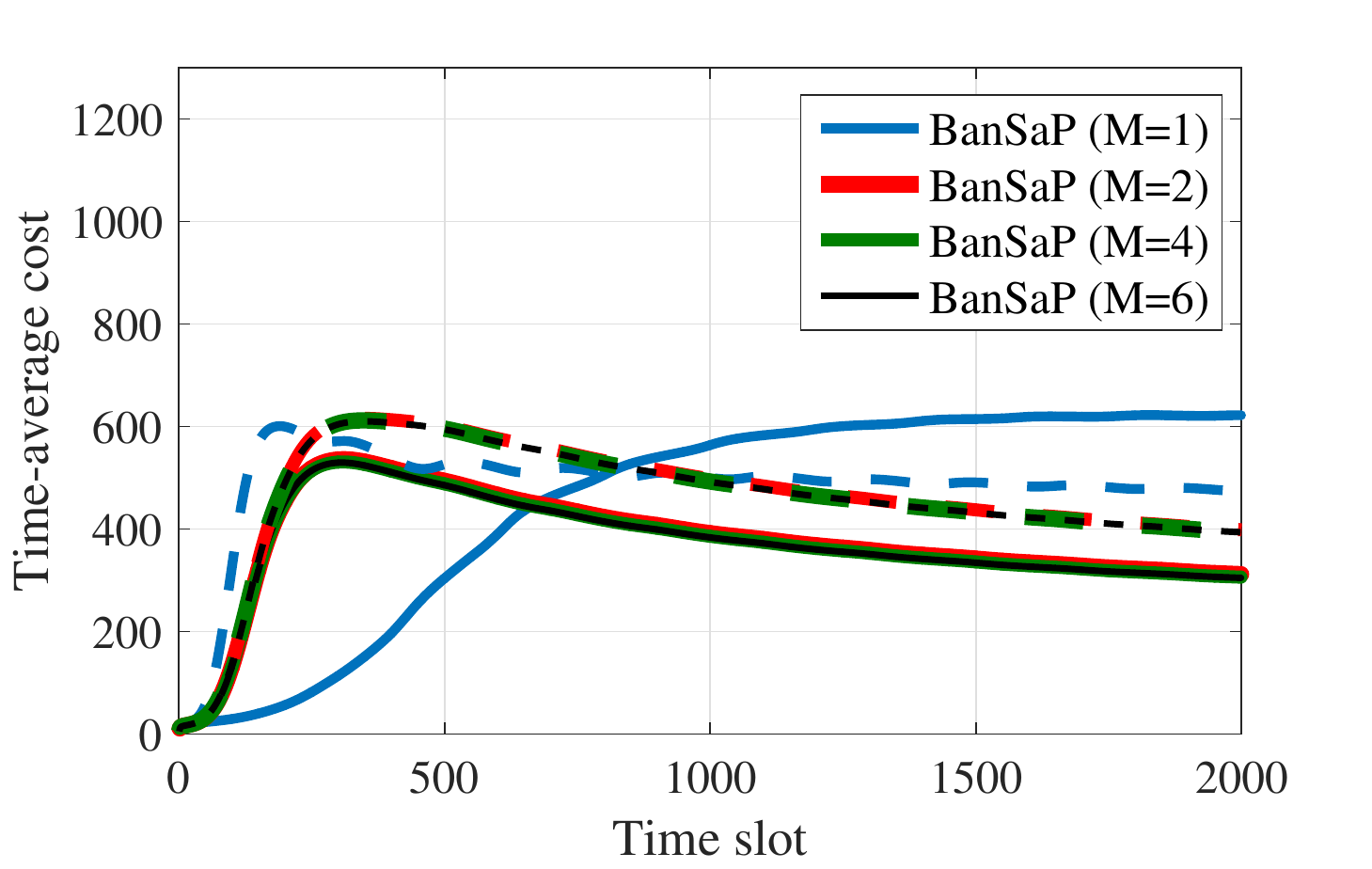}
\vspace{-0.2cm}
\caption{Effect of sampling schemes and number of feedback on average cost. Solid lines: BanSaP with sampling from a unit sphere (uniform sampling). Dashed lines: BanSaP with randomly sampling from standard basis (coordinate sampling).}
\label{Fig.reg1}
\end{figure}

Finally, 
BanSaP is benchmarked by: i) the \emph{full-information} MOSP method in \cite{chen2017tsp} that takes gradient-based update for primal-dual variables; ii) the \emph{heuristic} cloud-only approach that offloads all data requests to the remote cloud; and, iii) the \emph{heuristic} fog-only approach that processes all data requests locally without collaboration. 
For both cloud-only and fog-only approaches, unoffloaded and unprocessed requests are buffered at the fog nodes for later processing; thus, these amounts are measured by their fit.   
As different stepsizes of BanSaP and MOSP lead to different behaviors, we manually optimized stepsizes in each test so that they have similar fit, and focus on their cost comparison.
All simulated tests were averaged over 500 Monte Carlo realizations.

\emph{Effect of complexity and sampling schemes.}
In a simplified setting with $N=5$ nodes, the fit and average cost are compared among the BanSaP variants with $M$-point feedback under different sampling schemes in Figs. \ref{Fig.fit1} and \ref{Fig.reg1}. 
Clearly, for both sampling schemes, the cost and fit of BanSaP solvers decrease as the amount of bandit feedback increases.  
However, such performance gain varnishes when feedback increases; e.g., $M\geq 4$. 
Regarding the sampling schemes, 
Fig. \ref{Fig.fit1} demonstrates that when all the BanSaP variants have low dynamic fit, the uniform sampling-based BanSaP with one-point feedback has large initial fit; and Fig. \ref{Fig.reg1} confirms that  for $M=1$, the coordinate sampling-based BanSaP outperforms that with uniform sampling; and, for $M\geq 2$, the BanSaP solvers with uniform sampling incur lower cost.  
Therefore, to optimize empirical performance in the subsequent tests, coordinate sampling is adopted by BanSaP with $M=1$, while uniform sampling is used in BanSaP with $M\geq2$.

\begin{figure}[t]
\centering
\vspace{-0.4cm}
\hspace{-0.2cm}
\includegraphics[height=0.32\textwidth]{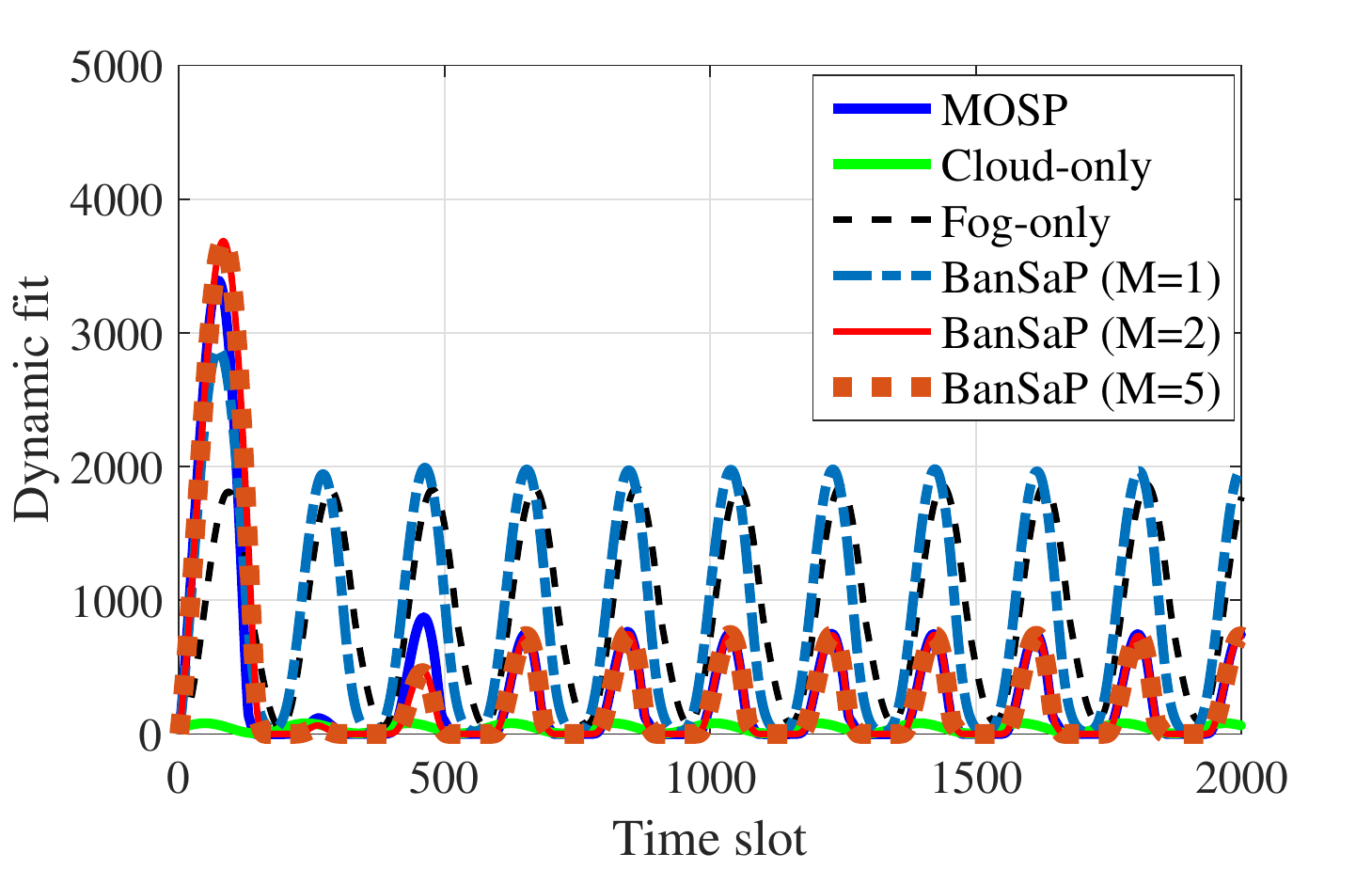}
\vspace{-0.2cm}
\caption{Comparison based on dynamic fit.}
\label{Fig.fit2}
\end{figure}

\begin{figure}[t]
\centering
\hspace{-0.2cm}
\vspace{-0.4cm}
\includegraphics[height=0.323\textwidth]{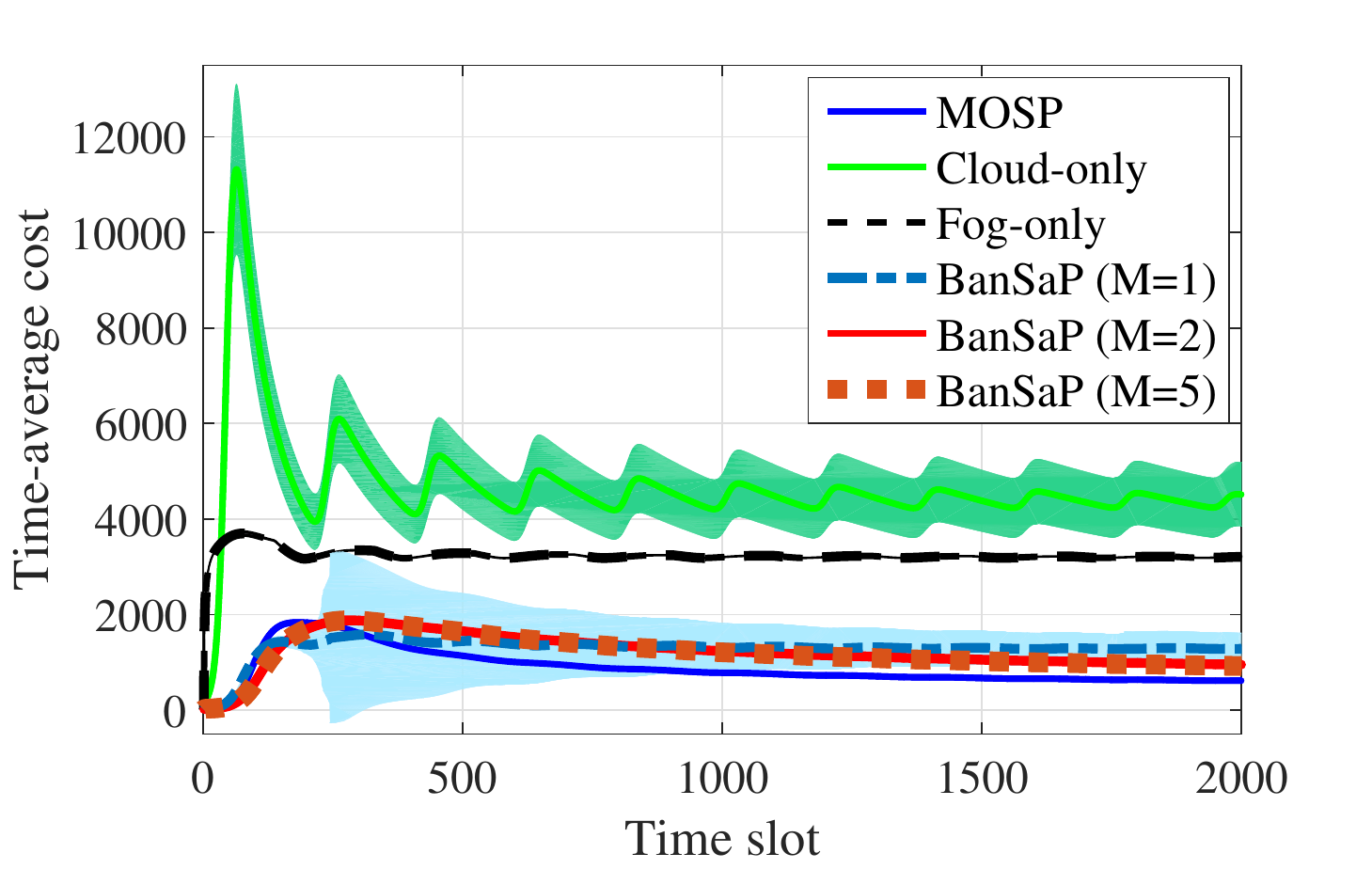}
\vspace{-0.1cm}
\caption{Comparison of average costs. The shaded region represents the cost distribution of each scheme within one standard deviation of the mean.}
\label{Fig.reg2}
\vspace{-0.1cm}
\end{figure}

\emph{Optimality and feasibility.}
With optimized sampling schemes for BanSaP solvers, the dynamic
fit and average cost are then compared among three BanSaP variants, MOSP, and two heuristic schemes in Figs. \ref{Fig.fit2} and \ref{Fig.reg2}. 
Without queueing at the fog side, the cloud-only scheme has much lower dynamic fit since all user demands are offloaded to the remote cloud. 
However, it incurs a much higher average cost (service latency) as the network latency between fog and cloud becomes high due to the large offloading amount.   
By increasing the amount of feedback, the BanSaP solver tends to have a lower fit and a lower average cost, both of which are comparable to those of MOSP when $M\geq 2$.
On the other hand, the BanSaP with only one-point bandit feedback still has a similar fit relative to the fog-only scheme, but enjoys much lower cost. 
Interestingly enough, when the variance (cf. the shaded area in Fig. \ref{Fig.reg2}) of the one-point BanSaP's cost is high, it markedly varnishes when multiple function values become available, which corroborates our claims in Theorems \ref{Them1}-\ref{Them2}. 

\begin{figure}[t]
\centering
\vspace{-0.4cm}
\hspace{-0.2cm}
\includegraphics[height=0.32\textwidth]{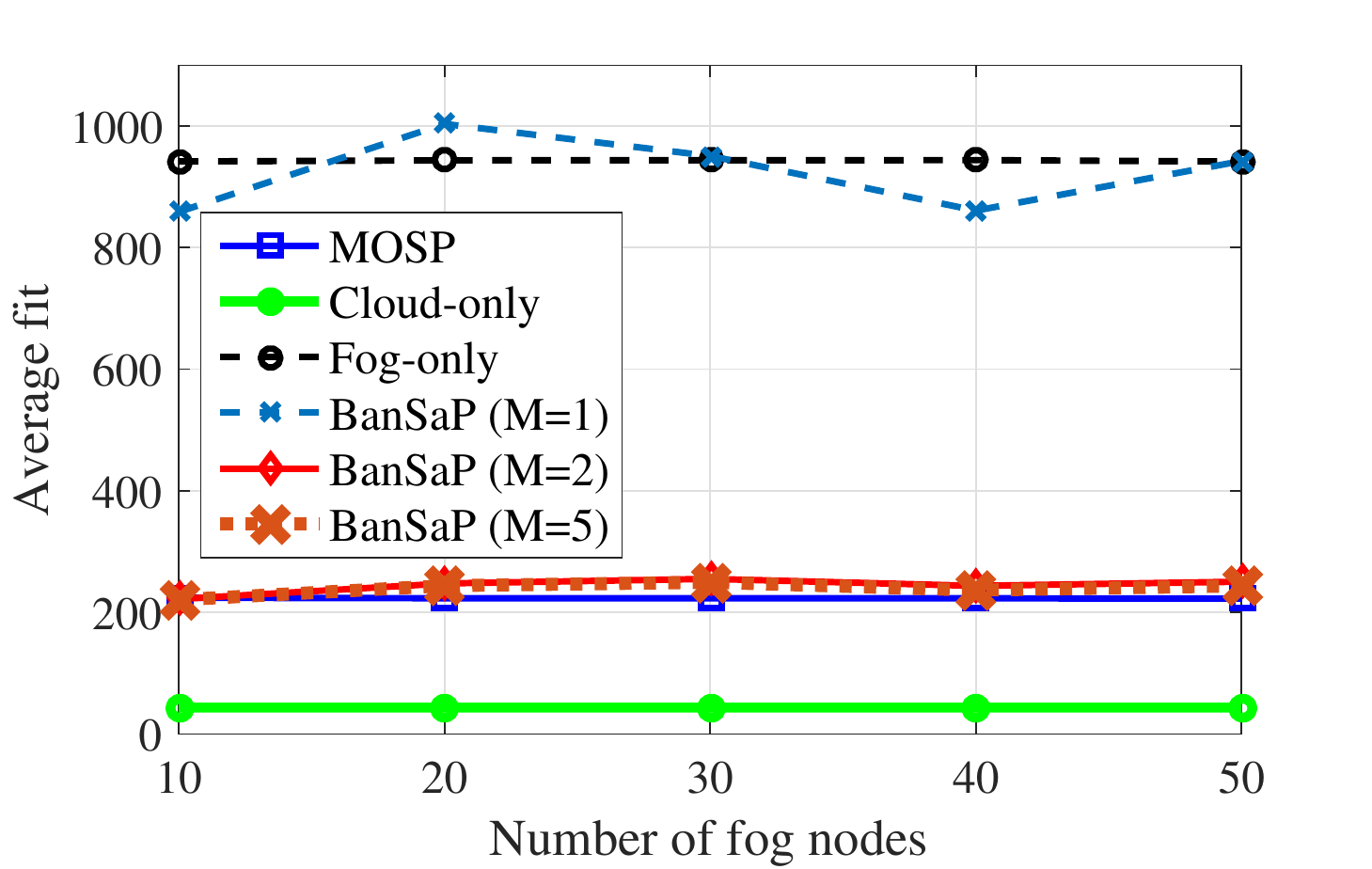}
\vspace{-0.2cm}
\caption{Impact of network size on dynamic fit per fog node.}
\label{Fig.fit3}
\end{figure}

\begin{figure}[t]
\centering
\hspace{-0.2cm}
\vspace{-0.3cm}
\includegraphics[height=0.32\textwidth]{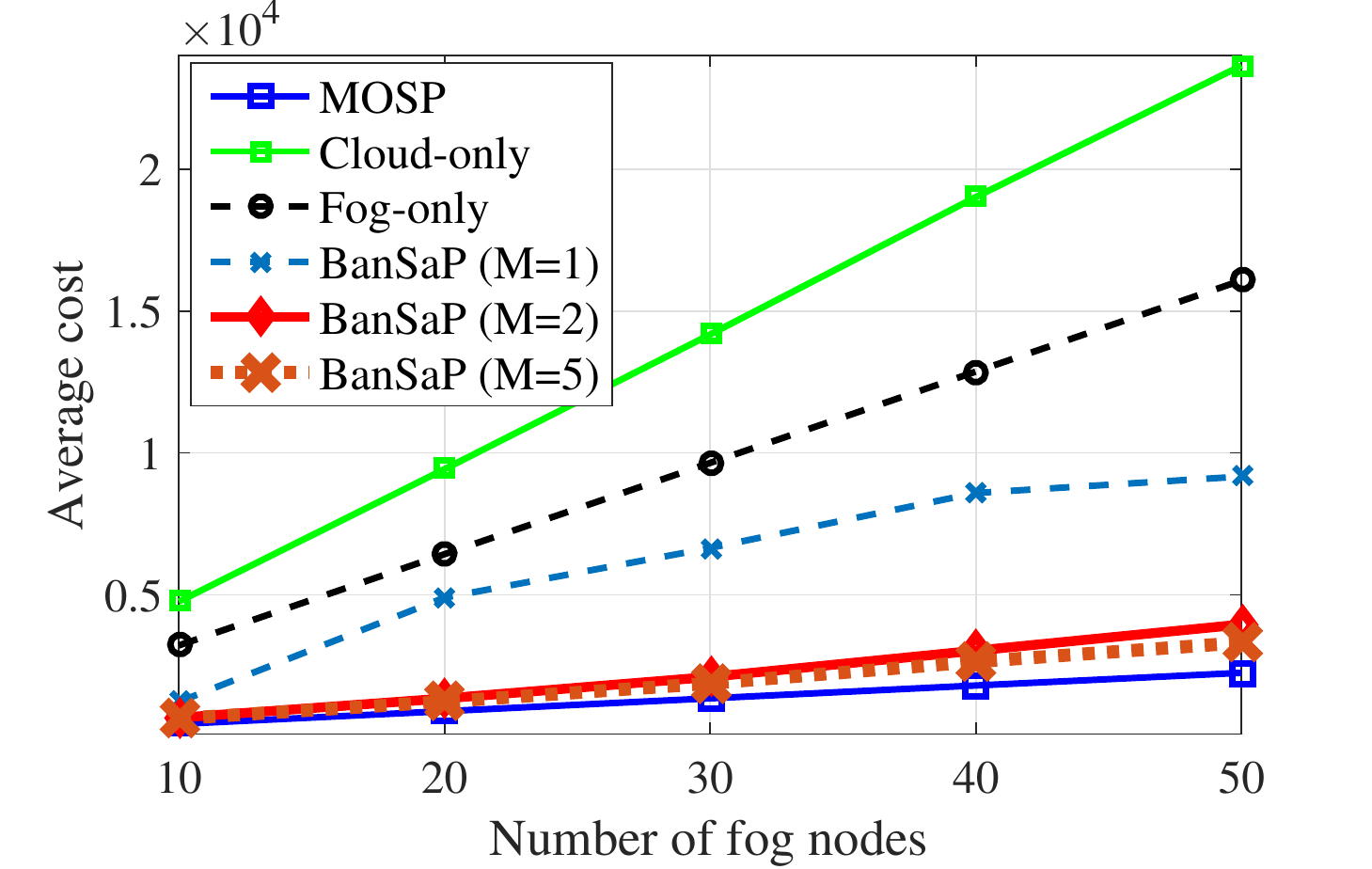}
\vspace{-0.2cm}
\caption{Impact of network size on average network cost.}
\label{Fig.reg3}
\end{figure}

\emph{Effect of network size.}
The third test evaluates the performance of all schemes under different number of fog nodes (i.e., network size).  
For each algorithm, the fit averaged over all fog nodes and time is presented in Fig. \ref{Fig.fit3}, and the cost averaged over the time is shown in Fig. \ref{Fig.reg3}. 
Clearly, the one-point BanSaP has lower average fit than the fog-only approach in most scenarios, and also incurs less average cost in all tested settings.  
Similar to those in Figs. \ref{Fig.fit2} and \ref{Fig.reg2}, the average fit of BanSaP with multiple function evaluations is still comparable to that of the full-information MOSP as the network size grows. 
An interesting observation here is that as the number of fog nodes increases, the performance gain of the BanSaP solver with a large $M$ becomes more evident; see e.g., Fig. \ref{Fig.reg3}. 
This implies that for a larger network, BanSaP benefits from more bandit information to learn and track the network dynamics.

\section{Conclusions and the Road Ahead}

Bandit convex optimization (BCO) in dynamic environments was studied
in this paper. 
Different from existing works in bandit settings, the focus was on a broader setting where part of the constraints are revealed after taking actions, and are also tolerable to instantaneous violations but have to be satisfied on average.  
The novel BCO setting fits well the emerging fog computing tasks in IoT.
A class of online bandit saddle-point (BanSaP) approaches were proposed, and their online performance was rigorously analyzed. 
It was shown that the resultant regret bounds match those attained in BCO setups without long-term constraints. 
Furthermore, the BanSaP solvers can simultaneously yield sub-linear dynamic regret and fit, if the dynamic solutions vary slowly over time. 

Our algorithmic and theoretical results serve as an exciting first step to \emph{innovate
online bandit learning tailored for dynamic network management tasks}, emerging from contemporary IoT applications. 
Interesting future directions include designing asynchronous variants of BanSaP, and incorporating predictable dynamic models in online network optimization. 

\section*{Acknowledgement}
The authors would like to thank Yanning Shen and Qing Ling for
helpful feedback on the early version of this manuscript.

\setlength{\belowdisplayskip}{4pt}
\setlength{\abovedisplayskip}{4pt}
\appendix
The proof generalizes the result in \cite{neely2017} from static regret with full-information gradient feedback to the dynamic regret with partial-information bandit feedback. 

\subsection{Supporting lemmas}
Before proving the main theory, we first establish several key lemmas and propositions.
The following lemma establishes the unbiasedness of one- and two-point estimations \cite{flaxman2005,agarwal2010}.

\begin{lemma}\label{lem.flaxman}
	With $\mathbf{u}$ drawn uniformly from the surface of the unit ball $\mathbb{S}:=\{\mathbf{u}:\|\mathbf{u}\|=1\}\subseteq \mathbb{R}^d$, we have for given a constant $\delta>0$ that
	\begin{equation}
           		\mathbb{E}_{\mathbf{u}}\left[\frac{d}{\delta}f_t(\mathbf{x}+\delta \mathbf{u})\mathbf{u}\right]=\nabla\check{f}_t(\mathbf{x})
	\end{equation}
where $\nabla\check{f}_t(\mathbf{x})$ is the gradient of the smoothed function $\check{f}_t(\mathbf{x}):=\mathbb{E}_{\mathbf{v}}[f_t(\mathbf{x}+\delta \mathbf{v})]$ with $\mathbf{v}$ drawn from a unit ball $\mathbb{B}$, and $d$ is the dimension of the variable $\mathbf{x}$.
Likewise, for the two-point case, we have that
	\begin{equation}
           		\mathbb{E}_{\mathbf{u}}\left[\frac{d}{2\delta}\Big(f_t(\mathbf{x}+\delta \mathbf{u})-f_t(\mathbf{x}-\delta \mathbf{u})\Big)\mathbf{u}\right]=\nabla\check{f}_t(\mathbf{x}).
	\end{equation}
\end{lemma}
Lemma \ref{lem.flaxman} provides valuable insights for performing gradient-based algorithms in bandit setting. Namely, $\hat{\nabla}^1 f_t(\hat{\mathbf{x}}_t)$ and $\hat{\nabla}^2 f_t(\hat{\mathbf{x}}_t)$ are the unbiased gradient estimators of the \emph{smoothed} function $\check{f}_t(\mathbf{x})$, which is an approximation of $f_t(\mathbf{x})$. Note that (as1)-(as2) also imply that the smoothed function $\check{f}_t(\mathbf{x})$ is convex and $G$-Lipschitz continuous \cite{agarwal2010}, which will be used frequently in the subsequent analysis.

The following lemma establishes the norm (or variance) of one- and two-point gradient estimations \cite{flaxman2005,agarwal2010}.

\begin{lemma}\label{app-lemma0}
For the gradient $\hat{\nabla}^1 f_t(\hat{\mathbf{x}}_t)$ in \eqref{eq.grad2}, we have that
\begin{align}\label{eq.grad-norm}
	\|\hat{\nabla}^1 f_t(\hat{\mathbf{x}}_t)\|\leq \frac{d}{\delta} F
\end{align}
where $F$ is an upper-bound of the function.  
For the gradient estimator $\hat{\nabla}^2 f_t(\hat{\mathbf{x}}_t)$ in \eqref{eq.grad3}, we have that
\begin{align}\label{eq.grad-norm2}
	\|\hat{\nabla}^2 f_t(\hat{\mathbf{x}}_t)\|\leq d G
\end{align}
where $G$ is the Lipschitz constant of the loss function.  
\end{lemma}
\begin{IEEEproof}
	For the gradient $\hat{\nabla}^1 f_t(\hat{\mathbf{x}}_t)$ in \eqref{eq.grad2}, it holds that
\begin{align}
	\|\hat{\nabla}^1 f_t(\hat{\mathbf{x}}_t)\|=\frac{d}{\delta}|f_t(\hat{\mathbf{x}}_t+\delta \mathbf{u}_t)|\|\mathbf{u}_t\|\leq \frac{d}{\delta} F
\end{align}
where $F$ is an upper-bound of the function. Likewise for $\hat{\nabla}^2 f_t(\hat{\mathbf{x}}_t)$ in \eqref{eq.grad3}, we have that
\begin{align}
	\|\hat{\nabla}^2 f_t(\hat{\mathbf{x}}_t)\|&=\Big\|\frac{d}{2\delta}\big(f_t(\hat{\mathbf{x}}_t+\delta \mathbf{u}_t)-f_t(\hat{\mathbf{x}}_t-\delta \mathbf{u}_t)\big)\mathbf{u}_t\Big\|\nonumber\\
	&=\frac{d}{2\delta}|f_t(\hat{\mathbf{x}}_t+\delta \mathbf{u}_t)-f_t(\hat{\mathbf{x}}_t-\delta \mathbf{u}_t)|\|\mathbf{u}_t\|\nonumber\\
	&\leq \frac{dG}{2\delta}\|2\delta \mathbf{u}_t\|
	\leq d G
\end{align}
where $G$ is the Lipschitz constant of the loss function and $\|\mathbf{u}_t\|=1$ by design. Thus, the proof is complete. 
\end{IEEEproof}

Having bounded the norm of stochastic gradients, the next lemma is useful to ensure feasibility of actual online actions.
\begin{lemma}[\!\!{\cite[Observation 2]{flaxman2005}}]\label{app-lemma1}
Consider a constant $r>0$ so that $r\mathbb{B}\subseteq {\cal X}$, where $\mathbb{B}:=\{\mathbf{v}:\|\mathbf{v}\|\leq 1\}\subseteq \mathbb{R}^d$ is the unit ball. If we choose $\gamma=\delta/r$, and the iterate satisfies $\hat{\mathbf{x}}_t\!\in\!(1-\gamma){\cal X}$, then $\hat{\mathbf{x}}_t+\delta \mathbf{u}_t\in{\cal X}$, where $\mathbf{u}_t$ is drawn uniformly from the unit sphere $\mathbb{S}:=\{\mathbf{u}:\|\mathbf{u}\|=1\}\subseteq \mathbb{R}^d$.
\end{lemma}

The next lemma is crucial to establish the dynamic fit \cite{neely2017}.
\begin{lemma}\label{app-lemma2}
	Considering the BanSaP recursion \eqref{eq.dual2}, we have the following bound for the cumulative constraint violation
\begin{align}\label{eq.prop1-1}
\sum_{t=1}^T \mathbf{g}_t(\hat{\mathbf{x}}_t)\leq \frac{\bm{\lambda}_{T+1}}{\mu}+\frac{G^2T\mathbf{1}}{2\beta}+\frac{\beta}{2}\sum_{t=1}^T\|\hat{\mathbf{x}}_{t+1}-\hat{\mathbf{x}}_t\|^2\mathbf{1}
\end{align}
where $\mu>0$ is the stepsize of the dual iteration \eqref{eq.dual}, and $\beta>0$ is a pre-defined constant. 
\end{lemma}

\begin{IEEEproof}
From the $n$-th entry of $\bm{\lambda}$ in \eqref{eq.dual2}, we have
\begin{align}\label{eq.app1-1}
\lambda_{t+1}^n&\geq \lambda_t^n+\mu (g_t^n(\hat{\mathbf{x}}_t)+\nabla^{\top} g_t^n(\hat{\mathbf{x}}_t)(\hat{\mathbf{x}}_{t+1}-\hat{\mathbf{x}}_t))\nonumber\\
&\!\stackrel{(a)}{\geq} \lambda_t^n+\mu g_t^n(\hat{\mathbf{x}}_t)-\frac{\mu}{2\beta}\|\nabla g_t^n(\hat{\mathbf{x}}_t)\|^2-\frac{\mu\beta}{2}\|\hat{\mathbf{x}}_{t+1}-\hat{\mathbf{x}}_t\|^2\nonumber\\
&\!\stackrel{(b)}{\geq} \lambda_t^n+\mu g_t^n(\hat{\mathbf{x}}_t)-\frac{\mu G^2}{2\beta}-\frac{\mu\beta}{2}\|\hat{\mathbf{x}}_{t+1}-\hat{\mathbf{x}}_t\|^2
\end{align}
where (a) uses the Cauchy-Schwarz inequality, and (b) follows from the bound on the gradients in (as2). The proof is then complete after summing up \eqref{eq.app1-1} over $t=1,\ldots,T$.
\end{IEEEproof}

\begin{lemma}\label{app-lemma3}
	Consider the BanSaP recursions \eqref{eq.primal2} and \eqref{eq.dual2} with a generic gradient $\hat{\nabla} f_t(\hat{\mathbf{x}}_t)$, which is estimated from one- or multi-point feedback. The following holds $\forall \mathbf{x}\in (1-\gamma){\cal X}$
\begin{align}\label{app-lemma3-1}
	\!\!\!\!&\frac{1}{\mu}\mathbb{E}[\Delta(\bm{\lambda}_t)]\leq \check{f}_t(\mathbf{x})-\mathbb{E}[\check{f}_t(\hat{\mathbf{x}}_t)]+\mathbb{E}[\bm{\lambda}_t^{\!\top}\mathbf{g}_t(\mathbf{x})]+\!2\mu G^2R^2\nonumber\\
\!\!\!\!&+\!\frac{1}{2\alpha}\mathbb{E}[\left\|\mathbf{x}\!-\!\hat{\mathbf{x}}_t\right\|^2]\!-\!\frac{1}{2\alpha}\mathbb{E}[\left\|\mathbf{x}\!-\!\hat{\mathbf{x}}_{t+1}\right\|^2]\!+\!\alpha\|\hat{\nabla} f_t(\hat{\mathbf{x}}_t)\|^2
\end{align}
where the constants $G$, $R$ and $F$ are as in (as2) and (as3).
\end{lemma}

\begin{IEEEproof}
Taking the norm square in \eqref{eq.dual2}, we have	
\begin{align}\label{eq.app1-2}
	\!\!\|\bm{\lambda}_{t+1}\|^2\!&\leq \|\bm{\lambda}_t\|^2+2\mu \bm{\lambda}_t^{\top}(\mathbf{g}_t(\hat{\mathbf{x}}_t)+\nabla^{\top} \mathbf{g}_t(\hat{\mathbf{x}}_t)(\hat{\mathbf{x}}_{t+1}-\hat{\mathbf{x}}_t))\nonumber\\
	&+\mu^2\|\mathbf{g}_t(\hat{\mathbf{x}}_t)+\nabla^{\top} \mathbf{g}_t(\hat{\mathbf{x}}_t)(\hat{\mathbf{x}}_{t+1}-\hat{\mathbf{x}}_t)\|^2\nonumber\\
	&\leq\|\bm{\lambda}_t\|^2+2\mu \bm{\lambda}_t^{\top}(\mathbf{g}_t(\hat{\mathbf{x}}_t)+\nabla^{\top} \mathbf{g}_t(\hat{\mathbf{x}}_t)(\hat{\mathbf{x}}_{t+1}-\hat{\mathbf{x}}_t))\nonumber\\
	&+\!2\mu^2\|\mathbf{g}_t(\hat{\mathbf{x}}_t)\|^2\!\!+\!2\mu^2\|\nabla^{\top} \mathbf{g}_t(\hat{\mathbf{x}}_t)(\hat{\mathbf{x}}_{t+1}\!-\!\hat{\mathbf{x}}_t)\|^2\!.\!
\end{align}

With $\Delta(\bm{\lambda}_t)\!:=\!\frac{1}{2}(\|\bm{\lambda}_{t+1}\|^2\!-\!\|\bm{\lambda}_t\|^2)$,  \eqref{eq.app1-2} implies that
\begin{equation}\label{eq.app1-3}
\frac{1}{\mu}\Delta(\bm{\lambda}_t)\!\leq\! \bm{\lambda}_t^{\!\top}\!(\mathbf{g}_t(\hat{\mathbf{x}}_t)\!+\!\nabla^{\top} \mathbf{g}_t(\hat{\mathbf{x}}_t)(\hat{\mathbf{x}}_{t+1}\!-\!\hat{\mathbf{x}}_t))\!+\!2\mu G^2R^2
\end{equation}
where $G$ and $R$ are the bounds on the gradient and the radius of the feasible set.
	
On the other hand, recall that the primal iterate $\hat{\mathbf{x}}_{t+1}$ is
the optimal solution to the following optimization problem
\begin{equation}\label{eq.app1-4}
	\hat{\mathbf{x}}_{t+1}\!=\!\arg\min_{\mathbf{x}\in(1-\gamma){\cal X}} \hat{\nabla}_{\mathbf{x}}^{\top}{\cal L}_t(\hat{\mathbf{x}}_t,\bm{\lambda}_t)(\mathbf{x}-\hat{\mathbf{x}}_t)+\frac{1}{2\alpha}\left\|\mathbf{x}-\hat{\mathbf{x}}_t\right\|^2.
\end{equation}
Recalling the definition of $\hat{\nabla}_{\mathbf{x}}{\cal L}_t(\hat{\mathbf{x}}_t,\bm{\lambda}_t)$, we thus have that
\begin{align}\label{eq.app1-5}
	\hat{\mathbf{x}}_{t+1}&\!=\!\arg\min_{\mathbf{x}\in(1-\gamma){\cal X}} \hat{\nabla}^{\top} f_t(\hat{\mathbf{x}}_t)(\mathbf{x}-\hat{\mathbf{x}}_t)\nonumber\\
	+&\bm{\lambda}_t^{\top}(\mathbf{g}_t(\hat{\mathbf{x}}_t)\!+\!\nabla^{\top} \mathbf{g}_t(\hat{\mathbf{x}}_t)(\mathbf{x}-\hat{\mathbf{x}}_t))+\frac{1}{2\alpha}\left\|\mathbf{x}-\hat{\mathbf{x}}_t\right\|^2
\end{align}
where we add $\bm{\lambda}_t^{\top}\mathbf{g}_t(\mathbf{x}_t)$ to the RHS of \eqref{eq.app1-4}. 
Note that it will not change the minimizer of \eqref{eq.app1-4}, since the added term is constant, and not coupled with the variable $\mathbf{x}$.

To connect \eqref{eq.app1-3} with the bound obtained in \eqref{eq.app1-5}, adding $\hat{\nabla}^{\top} f_t(\hat{\mathbf{x}}_t)(\hat{\mathbf{x}}_{t+1}-\hat{\mathbf{x}}_t)\!+\!\frac{1}{2\alpha}\left\|\hat{\mathbf{x}}_{t+1}-\hat{\mathbf{x}}_t\right\|^2$ to the RHS of \eqref{eq.app1-3}, we have that
\begin{align}\label{eq.app1-6}
&\frac{1}{\mu}\Delta(\bm{\lambda}_t)\!+\!\hat{\nabla}^{\top} f_t(\hat{\mathbf{x}}_t)(\hat{\mathbf{x}}_{t+1}-\hat{\mathbf{x}}_t)\!+\!\frac{1}{2\alpha}\left\|\hat{\mathbf{x}}_{t+1}\!-\!\hat{\mathbf{x}}_t\right\|^2\nonumber\\
\leq &\, \bm{\lambda}_t^{\!\top}\left(\mathbf{g}_t(\hat{\mathbf{x}}_t)\!+\!\nabla^{\top} \mathbf{g}_t(\hat{\mathbf{x}}_t)(\hat{\mathbf{x}}_{t+1}\!-\!\hat{\mathbf{x}}_t)\right)\!+\!\frac{1}{2\alpha}\left\|\hat{\mathbf{x}}_{t+1}-\hat{\mathbf{x}}_t\right\|^2\nonumber\\
	\!&\quad\quad~+ \hat{\nabla}^{\top} f_t(\hat{\mathbf{x}}_t)\left(\hat{\mathbf{x}}_{t+1}-\hat{\mathbf{x}}_t\right)\!+\!2\mu G^2R^2.
\end{align}

Note that $\hat{\mathbf{x}}_{t+1}$ is the minimizer of \eqref{eq.app1-5}, where the objective on the RHS of \eqref{eq.app1-5} is strongly-convex, thus we have that
\begin{align}\label{eq.app1-7}
	&\frac{1}{\mu}\Delta(\bm{\lambda}_t)+\hat{\nabla}^{\top} f_t(\hat{\mathbf{x}}_t)(\hat{\mathbf{x}}_{t+1}\!-\!\hat{\mathbf{x}}_t)\!+\!\frac{1}{2\alpha}\left\|\hat{\mathbf{x}}_{t+1}\!-\!\hat{\mathbf{x}}_t\right\|^2\nonumber\\
	\!\leq &\,\bm{\lambda}_t^{\!\top}\left(\mathbf{g}_t(\hat{\mathbf{x}}_t)\!+\!\nabla^{\top}\! \mathbf{g}_t(\hat{\mathbf{x}}_t)(\mathbf{x}-\!\hat{\mathbf{x}}_t)\right)\!+\!\frac{1}{2\alpha}\left\|\mathbf{x}-\hat{\mathbf{x}}_t\right\|^2\!\!+\!2\mu G^2R^2\nonumber\\
	&+\hat{\nabla}^{\!\top} f_t(\hat{\mathbf{x}}_t)\!\left(\mathbf{x}-\hat{\mathbf{x}}_t\right)\!-\!\frac{1}{2\alpha}\left\|\mathbf{x}-\hat{\mathbf{x}}_{t+1}\right\|^2\nonumber\\
\!\!\!\stackrel{(a)}{\leq} &\,\bm{\lambda}_t^{\!\top}\mathbf{g}_t(\mathbf{x})+\hat{\nabla}^{\top} f_t(\hat{\mathbf{x}}_t)\!\left(\mathbf{x}-\hat{\mathbf{x}}_t\right)\!+\!2\mu G^2R^2\nonumber\\
	&+\!\frac{1}{2\alpha}\left\|\mathbf{x}-\hat{\mathbf{x}}_t\right\|^2\!-\!\frac{1}{2\alpha}\left\|\mathbf{x}-\hat{\mathbf{x}}_{t+1}\right\|^2\!\!,~\forall \mathbf{x}\in (1-\gamma){\cal X}
\end{align}
where (a) uses the non-negativity that $\bm{\lambda}_t\geq \mathbf{0}$, and the convexity such that $\mathbf{g}_t(\hat{\mathbf{x}}_t)\!+\!\nabla^{\!\top} \mathbf{g}_t(\hat{\mathbf{x}}_t)(\mathbf{x}-\hat{\mathbf{x}}_t)\leq \mathbf{g}_t(\mathbf{x})$.

Using the Cauchy-Schwarz inequality, we have that
\begin{align}\label{eq.app1-8}
	\!\!-\hat{\nabla}^{\top} f_t(\hat{\mathbf{x}}_t)(\hat{\mathbf{x}}_{t+1}\!-\!\hat{\mathbf{x}}_t)
	\!\leq\! \alpha\|\hat{\nabla} f_t(\hat{\mathbf{x}}_t)\|^2+\frac{\|\hat{\mathbf{x}}_{t+1}\!-\!\hat{\mathbf{x}}_t\|^2}{4\alpha}.
\end{align}

Plugging \eqref{eq.app1-8} into \eqref{eq.app1-7} and rearranging terms, for $\forall \mathbf{x}\in (1-\gamma){\cal X}$, we have that 
 \begin{align}\label{eq.app1-9}
	&\frac{1}{\mu}\Delta(\bm{\lambda}_t)\!+\!\frac{1}{4\alpha}\left\|\hat{\mathbf{x}}_{t+1}\!-\!\hat{\mathbf{x}}_t\right\|^2\nonumber\\
	\leq &\,\bm{\lambda}_t^{\!\top}\mathbf{g}_t(\mathbf{x})+\hat{\nabla}^{\top} f_t(\hat{\mathbf{x}}_t)\left(\mathbf{x}-\hat{\mathbf{x}}_t\right)\!+\!2\mu G^2R^2\nonumber\\
	&+\!\frac{1}{2\alpha}\left\|\mathbf{x}-\hat{\mathbf{x}}_t\right\|^2\!-\!\frac{1}{2\alpha}\left\|\mathbf{x}-\hat{\mathbf{x}}_{t+1}\right\|^2+\alpha\|\hat{\nabla} f_t(\hat{\mathbf{x}}_t)\|^2.
\end{align}

Taking expectation over $\mathbf{u}_t$ on both side of \eqref{eq.app1-9} conditioning on $\hat{\mathbf{x}}_t$, it follows that
 \begin{align}\label{eq.app1-10}
	&\frac{1}{\mu}\mathbb{E}[\Delta(\bm{\lambda}_t)]\!+\!\frac{1}{4\alpha}\mathbb{E}[\left\|\hat{\mathbf{x}}_{t+1}\!-\!\hat{\mathbf{x}}_t\right\|^2]\nonumber\\
	\!\!\!\!\leq &\,\bm{\lambda}_t^{\!\top}\mathbf{g}_t(\mathbf{x})+\mathbb{E}\left[\hat{\nabla}^{\top} f_t(\hat{\mathbf{x}}_t)\left(\mathbf{x}-\hat{\mathbf{x}}_t\right)\right]\!+\!2\mu G^2R^2\nonumber\\
	&+\frac{1}{2\alpha}\left\|\mathbf{x}-\hat{\mathbf{x}}_t\right\|^2\!-\!\frac{1}{2\alpha}\mathbb{E}[\left\|\mathbf{x}-\hat{\mathbf{x}}_{t+1}\right\|^2]+\alpha\|\hat{\nabla} f_t(\hat{\mathbf{x}}_t)\|^2\nonumber\\
\!\!\!\!\stackrel{(c)}{=}&\,\bm{\lambda}_t^{\!\top}\mathbf{g}_t(\mathbf{x})+\nabla^{\top}\check{f}_t(\hat{\mathbf{x}}_t)\left(\mathbf{x}-\hat{\mathbf{x}}_t\right)\!+\!2\mu G^2R^2\nonumber\\
	&+\frac{1}{2\alpha}\left\|\mathbf{x}-\hat{\mathbf{x}}_t\right\|^2\!-\!\frac{1}{2\alpha}\mathbb{E}[\left\|\mathbf{x}-\hat{\mathbf{x}}_{t+1}\right\|^2]+\alpha\|\hat{\nabla} f_t(\hat{\mathbf{x}}_t)\|^2
\end{align}
where (c) holds since the randomness $\mathbf{u}_t$ in $\hat{\nabla} f_t(\hat{\mathbf{x}}_t)$ is independent of $\hat{\mathbf{x}}_t$, and $\hat{\nabla} f_t(\hat{\mathbf{x}}_t)$ is an unbiased estimator of $\nabla\check{f}_t(\hat{\mathbf{x}}_t)$.

The convexity of $f_t(\mathbf{x})$ implies that $\check{f}_t(\mathbf{x})$ is also convex, and thus $\nabla^{\top}\check{f}_t(\hat{\mathbf{x}}_t)\left(\mathbf{x}-\hat{\mathbf{x}}_t\right)\leq \check{f}_t(\mathbf{x})-\check{f}_t(\hat{\mathbf{x}}_t)$. Plugging into \eqref{eq.app1-10} and taking expectation over all possible $\hat{\mathbf{x}}_t$, it follows that
 \begin{align}\label{eq.app1-11}
 		\!&\frac{1}{\mu}\mathbb{E}[\Delta(\bm{\lambda}_t)]\!+\!\frac{1}{4\alpha}\mathbb{E}[\left\|\hat{\mathbf{x}}_{t+1}\!-\!\hat{\mathbf{x}}_t\right\|^2]\\
 	\!\!\leq &\,\check{f}_t(\mathbf{x})-\mathbb{E}[\check{f}_t(\hat{\mathbf{x}}_t)]+\mathbb{E}[\bm{\lambda}_t^{\!\top}\mathbf{g}_t(\mathbf{x})]+\!2\mu G^2R^2\nonumber\\
	\!\!+&\frac{1}{2\alpha}\mathbb{E}[\left\|\mathbf{x}-\hat{\mathbf{x}}_t\right\|^2]\!-\!\frac{1}{2\alpha}\mathbb{E}[\left\|\mathbf{x}-\hat{\mathbf{x}}_{t+1}\right\|^2]\!+\!\alpha\mathbb{E}[\|\hat{\nabla} f_t(\hat{\mathbf{x}}_t)\|^2]\!\nonumber
 \end{align}
 which completes the proof by dropping the nonnegative term $\mathbb{E}[\left\|\hat{\mathbf{x}}_{t+1}\!-\!\hat{\mathbf{x}}_t\right\|^2]$ in the LHS.
 \end{IEEEproof}
 
\subsection{Proof of Theorem \ref{Them1}}\label{app-prf1}
With $\gamma=\delta/r$, the feasibility of actions $\{\mathbf{x}_{1,t}\}$ readily follows from Lemma \ref{app-lemma1}, i.e., $\mathbf{x}_{1,t}\in{\cal X},\,\forall t$. 
To prove the dynamic regret and fit bounds, the following result is needed.
 \begin{lemma}\label{app-lemma4}
	For the BanSaP recursions \eqref{eq.primal2}-\eqref{eq.dual2}, if we choose $\alpha=\mu={\cal O}(T^{-\frac{3}{4}})$ and $\delta={\cal O}(T^{-\frac{1}{4}})$, the dual iterates are uniformly bounded by $\|\bm{\lambda}_t\|\leq C={\cal O}(1)$,
with the constant $C$ given by
\begin{equation}\label{eq.defi-C}
\!\!	C\!:=\!\max\!\left\{\!2GR,\Big(\frac{1}{\eta}\!+\!1\!\Big)GR\!+\!\frac{2G^2R^2\mu}{\eta}\!+\!\frac{d^2F^2\alpha}{\eta\delta^2}\!+\!\frac{\mu R^2}{2\alpha\eta}\!\right\}\!\!\!\!\!
\end{equation}
where the constants $G$, $R$, and $\eta$ are as in (as2)-(as4).
\end{lemma}

\begin{IEEEproof}
Plugging the bounded norm of the one-point gradient estimator \eqref{eq.grad-norm} into \eqref{app-lemma3-1}, it holds that 
 \begin{align}\label{eq.app1-15}
 		\frac{1}{\mu}\mathbb{E}[\Delta(\bm{\lambda}_t)]\leq &\,GR+\mathbb{E}[\bm{\lambda}_t^{\!\top}\mathbf{g}_t(\mathbf{x})]+\!2\mu G^2R^2\!+\!\frac{d^2 F^2\alpha}{\delta^2}\nonumber\\
	&\!+\!\frac{1}{2\alpha}\mathbb{E}[\left\|\mathbf{x}-\hat{\mathbf{x}}_t\right\|^2]\!-\!\frac{1}{2\alpha}\mathbb{E}[\left\|\mathbf{x}-\hat{\mathbf{x}}_{t+1}\right\|^2]
 \end{align}
 where we used the Lipschitz condition on \eqref{app-lemma3-1}; i.e., 
 \begin{equation}
 	 \mathbb{E}[\check{f}_t(\mathbf{x})-\check{f}_t(\hat{\mathbf{x}}_t)]\leq GR.
 \end{equation}
 
 Selecting the interior point $\mathbf{x}=\tilde{\mathbf{x}}\in (1-\gamma){\cal X}$ so that $\mathbf{g}_t(\tilde{\mathbf{x}})\leq -\eta\mathbf{1}$, it follows from \eqref{eq.app1-15} that
  \begin{align}\label{eq.app1-16}
  	 		\frac{1}{\mu}\mathbb{E}[\Delta(\bm{\lambda}_t)]\leq &\,GR-\eta\mathbb{E}[\bm{\lambda}_t^{\top}\mathbf{1}]+\!2\mu G^2R^2\!+\!\frac{d^2F^2\alpha}{\delta^2}\nonumber\\
	&\!+\!\frac{1}{2\alpha}\mathbb{E}[\left\|\tilde{\mathbf{x}}-\hat{\mathbf{x}}_t\right\|^2]\!-\!\frac{1}{2\alpha}\mathbb{E}[\left\|\tilde{\mathbf{x}}-\hat{\mathbf{x}}_{t+1}\right\|^2].
  \end{align}
  
Using $-\eta\bm{\lambda}_t^{\!\top}\mathbf{1}=-\eta\|\bm{\lambda}_t\|_1\leq -\eta\|\bm{\lambda}_t\|$, we arrive at
  \begin{align}\label{eq.app1-17}
  	 		\frac{1}{\mu}\mathbb{E}[\Delta(\bm{\lambda}_t)]\leq &\,GR-\eta\mathbb{E}[\|\bm{\lambda}_t\|]+\!2\mu G^2R^2\!+\!\frac{d^2F^2\alpha}{\delta^2}\nonumber\\
	&\!+\!\frac{1}{2\alpha}\mathbb{E}[\left\|\tilde{\mathbf{x}}-\hat{\mathbf{x}}_t\right\|^2]\!-\!\frac{1}{2\alpha}\mathbb{E}[\left\|\tilde{\mathbf{x}}-\hat{\mathbf{x}}_{t+1}\right\|^2].
  \end{align}

Now we are ready to show that the norm of the dual variable is uniformly bounded by a constant $C$ that is independent of time; that is, $\|\bm{\lambda}_t\|\leq C,\;\forall t$.

For $1\leq t\leq \frac{1}{\mu}$, it follows readily that
\begin{align}\label{eq.app1-18a}
	\|\bm{\lambda}_t\|&\leq \|\bm{\lambda}_{t-1}\|+\mu \|\mathbf{g}_t(\hat{\mathbf{x}}_t)+\nabla^{\top} \mathbf{g}_t(\hat{\mathbf{x}}_t)(\hat{\mathbf{x}}_{t+1}-\hat{\mathbf{x}}_t)\|\nonumber\\
	&\leq \|\bm{\lambda}_{t-1}\|+2\mu GR\leq \|\bm{\lambda}_1\|+2\mu tGR\leq C
\end{align}
where the last inequality follows from $\bm{\lambda}_1=\mathbf{0}$, $t\leq {1}/{\mu}$, and the definition of $C$ in \eqref{eq.defi-C}. 

For $\frac{1}{\mu}\leq t\leq T$, we will prove the claim by contradiction. Assume $T_0$ is the first slot for which $\|\bm{\lambda}_{T_0}\|>C$. Therefore, we have $\|\bm{\lambda}_{T_0}\|>C\geq \|\bm{\lambda}_{T_0-\frac{1}{\mu}}\|$, which after recalling \eqref{eq.app1-17} and the definition of $\Delta(\bm{\lambda}_t)$, yields
\begin{equation}\label{eq.app1-18b}
	\frac{1}{\mu}\!\!\sum_{t=T_0-\frac{1}{\mu}}^{T_0-1}\!\!\mathbb{E}[\Delta(\bm{\lambda}_t)]\!=\!\frac{1}{2\mu}\left(\mathbb{E}\Big[\|\bm{\lambda}_{T_0}\|^2\!-\!\|\bm{\lambda}_{T_0-\frac{1}{\mu}}\|^2\Big]\right)\!>\!0.
\end{equation}
On the other hand however, summing up \eqref{eq.app1-17}, we obtain 
\begin{align}\label{eq.app1-18c}
	&\frac{1}{\mu}\!\sum_{t=T_0-\frac{1}{\mu}}^{T_0-1}\!\mathbb{E}[\Delta(\bm{\lambda}_t)]\leq \frac{GR}{\mu}-\eta\!\!\!\sum_{t=T_0-\frac{1}{\mu}}^{T_0-1}\!\!\mathbb{E}[\|\bm{\lambda}_t\|]+\!2G^2R^2\nonumber\\
	&~~~+\!\frac{d^2F^2\alpha}{\mu\delta^2}+\frac{1}{2\alpha}\mathbb{E}[\|\tilde{\mathbf{x}}-\hat{\mathbf{x}}_{T_0-\frac{1}{\mu}}\|^2]\!-\!\frac{1}{2\alpha}\mathbb{E}[\left\|\tilde{\mathbf{x}}-\hat{\mathbf{x}}_{T_0}\right\|^2]\nonumber\\
\stackrel{(a)}{\leq}  &\frac{GR}{\mu}-\eta\!\!\!\sum_{t=T_0-\frac{1}{\mu}}^{T_0-1}\!\!\!\mathbb{E}[\|\bm{\lambda}_t\|]+\!2G^2R^2\!+\!\frac{d^2F^2\alpha}{\mu\delta^2}+\frac{R^2}{2\alpha}	
\end{align}
where (a) uses again the bound $\|\tilde{\mathbf{x}}-\hat{\mathbf{x}}_{T_0-\frac{1}{\mu}}\|\leq R$.

Note that since $\|\bm{\lambda}_{T_0}\|>C$ and $\|\bm{\lambda}_{T_0}\|-\|\bm{\lambda}_{T_0-1}\|\leq 2\mu GR$, we have that
\begin{equation}\label{eq.app1-18d}
	\|\bm{\lambda}_{T_0-\tau}\|>C-2\tau \mu GR.
\end{equation}

Combining \eqref{eq.app1-18c} with \eqref{eq.app1-18d}, we deduce
\begin{align}\label{eq.app1-18e}
\!\frac{1}{\mu}\!\!\!\sum_{t=T_0-\frac{1}{\mu}}^{T_0-1}\!\!\!\mathbb{E}[\Delta(\bm{\lambda}_t)]\!\leq\! \frac{GR}{\mu}\!-\!\frac{C\eta}{\mu}\!+\!\frac{\eta GR}{\mu}\!+\!2G^2R^2\!+\!\frac{d^2F^2\alpha}{\mu\delta^2}+\!\frac{R^2}{2\alpha}.
\end{align}

Together with \eqref{eq.app1-18b}, recursion \eqref{eq.app1-18e} implies that
\begin{equation}
	C<\frac{GR}{\eta}+GR+\frac{2G^2R^2\mu}{\eta}+\frac{d^2F^2\alpha}{\eta\delta^2}+\!\frac{\mu R^2}{2\alpha\eta}
\end{equation}
which contradicts the definition of $C$ in \eqref{eq.defi-C}. 
Hence, there is no $T_0$ satisfying $\|\bm{\lambda}_t\|\leq C$, which implies that $\|\bm{\lambda}_t\|\leq C,\;\forall t$.


By choosing the stepsizes $\alpha=\mu={\cal O}(T^{-\frac{3}{4}})$, and the parameter $\delta={\cal O}(T^{-\frac{1}{4}})$, it follows that
\begin{equation}\label{eq.order-C}
	C\!=\!{\cal O}\!\left(\frac{GR}{\eta}\!+\!GR\!+\!\frac{2G^2R^2}{\eta T^{\frac{3}{4}}}\!+\!\frac{d^2F^2}{\eta T^{\frac{1}{4}}}\!+\!\frac{R^2}{2\eta}\right)\!=\!{\cal O}(1) 
\end{equation}	
which completes the proof of the lemma.
\end{IEEEproof}

\textbf{Dynamic regret in Theorem 1:}
 Recall that $\mathbf{x}_t^*$ is the minimizer of the following time-varying problem \eqref{eq.realtime-prob}, and note that $(1-\gamma)\mathbf{x}_t^*\in (1-\gamma){\cal X}$.
Hence, plugging $(1-\gamma)\mathbf{x}_t^*$ into \eqref{app-lemma3-1}, we have 
  \begin{align}\label{eq.app2-1}
	&\frac{1}{\mu}\mathbb{E}[\Delta(\bm{\lambda}_t)]\leq \check{f}_t((1-\gamma)\mathbf{x}_t^*)-\mathbb{E}[\check{f}_t(\hat{\mathbf{x}}_t)]\nonumber\\
+&\frac{1}{2\alpha}\mathbb{E}[\left\|(1-\gamma)\mathbf{x}_t^*-\hat{\mathbf{x}}_t\right\|^2]\!-\!\frac{1}{2\alpha}\mathbb{E}[\left\|(1-\gamma)\mathbf{x}_t^*-\hat{\mathbf{x}}_{t+1}\right\|^2]\nonumber\\
+&\mathbb{E}[\bm{\lambda}_t^{\!\top}\mathbf{g}_t((1-\gamma)\mathbf{x}_t^*)]+\frac{\alpha}{\delta^2}d^2F^2+2\mu G^2R^2.
 \end{align}
 
From the Lipschitz condition, we can bound the inner product in \eqref{eq.app2-1} by 
\begin{align}\label{eq.app2-2}
	&\mathbb{E}[\bm{\lambda}_t^{\!\top}\mathbf{g}_t((1-\gamma)\mathbf{x}_t^*)]\nonumber\\
	\leq &\mathbb{E}[\bm{\lambda}_t^{\!\top}(\mathbf{g}_t(\mathbf{x}_t^*)\!+\!\gamma GR\!\cdot\!\mathbf{1})]\stackrel{(a)}{\leq} \gamma GR \mathbb{E}[\|\bm{\lambda}_t\|]\!\stackrel{(b)}{\leq}\! \gamma GR\|\bar{\bm{\lambda}}\|
\end{align}
where (a) follows from $\bm{\lambda}_t^{\top}\mathbf{g}_t(\mathbf{x}_t^*)\leq 0$ since $\mathbf{g}_t(\mathbf{x}_t^*)\leq \mathbf{0}$, and $\bm{\lambda}_t\geq \mathbf{0}$; and (b) uses the upper bound of $\|\bar{\bm{\lambda}}\|:=\max_t\|\bm{\lambda}_t\|$.
The two distance terms in \eqref{eq.app2-1} can be bounded by
   \begin{align}\label{eq.app2-3}
        \|(1-\gamma)\mathbf{x}^*_t&-\!\hat{\mathbf{x}}_t\|^2\!-\|(1-\gamma)\mathbf{x}^*_t\!-\!\hat{\mathbf{x}}_{t+1}\|^2\nonumber\\
       = \|(1-\gamma)\mathbf{x}^*_t&-\!\hat{\mathbf{x}}_t\|^2\!-\|(1-\gamma)\mathbf{x}^*_{t-1}-\hat{\mathbf{x}}_t\|^2\nonumber\\
       &+\|(1-\gamma)\mathbf{x}^*_{t-1}-\hat{\mathbf{x}}_t\|^2\!-\|(1-\gamma)\mathbf{x}^*_t\!-\!\hat{\mathbf{x}}_{t+1}\|^2\nonumber\\
        = (1-\gamma)\|\mathbf{x}^*_t&-\mathbf{x}^*_{t-1}\|\|(1-\gamma)(\mathbf{x}^*_t+\mathbf{x}^*_{t-1})-2\hat{\mathbf{x}}_t\|\nonumber\\
        +\|(1&-\gamma)\mathbf{x}^*_{t-1}-\hat{\mathbf{x}}_t\|^2-\|(1-\gamma)\mathbf{x}^*_t\!-\!\hat{\mathbf{x}}_{t+1}\|^2\!.\!
   \end{align}

   Using the triangle inequality, it follows that
   \begin{align}
  & \|(1\!-\gamma)(\mathbf{x}^*_t+\mathbf{x}^*_{t-1})-2\hat{\mathbf{x}}_t\|\nonumber\\
   \leq & \|(1\!-\gamma)\mathbf{x}^*_t-\hat{\mathbf{x}}_t\|+\|(1-\gamma)\mathbf{x}^*_{t-1}-\hat{\mathbf{x}}_t\|\leq 2R	
   \end{align}
which together with \eqref{eq.app2-3}, implies that
   \begin{align}\label{eq.app2-3b}
   	 \|(1-\gamma)\mathbf{x}^*_t&-\!\hat{\mathbf{x}}_t\|^2\!-\|(1-\gamma)\mathbf{x}^*_t\!-\!\hat{\mathbf{x}}_{t+1}\|^2\nonumber\\
   	   \leq2(1 -\gamma)R&\|\mathbf{x}^*_t-\mathbf{x}^*_{t-1}\|+\|(1-\gamma)\mathbf{x}^*_{t-1}-\hat{\mathbf{x}}_t\|^2\nonumber\\
       &\qquad\qquad\quad~-\|(1-\gamma)\mathbf{x}^*_t\!-\!\hat{\mathbf{x}}_{t+1}\|^2.
   \end{align}

 Plugging \eqref{eq.app2-2} and \eqref{eq.app2-3b} into \eqref{eq.app2-1}, and summing up over $t=1,\ldots,T$, we find
  \begin{align}\label{eq.app1-12}
 		&\frac{1}{2\mu}\left(\mathbb{E}[\|\bm{\lambda}_{T+1}\|^2\!-\!\|\bm{\lambda}_1\|^2]\right)\!+\!\sum_{t=1}^T\!\left(\mathbb{E}[\check{f}_t(\hat{\mathbf{x}}_t)]\!-\!\check{f}_t((1-\gamma)\mathbf{x}^*_t)\right)\nonumber\\
 	\!\!\!\leq &\gamma GR\|\bar{\bm{\lambda}}\|T\!+\!\sum_{t=1}^T\frac{(1\!-\!\gamma)R}{\alpha}\|\mathbf{x}^*_t\!-\!\mathbf{x}^*_{t-1}\|\!+\!2\mu G^2R^2T\!+\!\frac{\alpha d^2F^2 T}{\delta^2}\nonumber\\
	\!\!\!+&\frac{1}{2\alpha}\Big(\mathbb{E}\big[\left\|(1-\gamma)\mathbf{x}^*_0-\hat{\mathbf{x}}_1\right\|^2\big]\!-\!\mathbb{E}\big[\left\|(1-\gamma)\mathbf{x}^*_T-\hat{\mathbf{x}}_{T+1}\right\|^2\big]\Big)\nonumber\\
	\!\!\!\!\stackrel{(c)}{\leq} &\,\gamma GR\|\bar{\bm{\lambda}}\|T\!+\!\frac{R}{\alpha}V(\mathbf{x}_{1:T}^*)\!+2\mu G^2R^2T\!+\frac{R^2}{2\alpha}\!+\!\frac{\alpha d^2F^2 T}{\delta^2}\!
 \end{align}
 where (c) uses $\|(1-\gamma)\mathbf{x}^*_0-\hat{\mathbf{x}}_1\|\leq \left\|\mathbf{x}^*_0-\hat{\mathbf{x}}_1\right\|\leq R$, and $\|(1-\gamma)\mathbf{x}^*_T-\hat{\mathbf{x}}_{T+1}\|^2\geq 0$, and the accumulated variation of the per-slot minimizers defined as $ V(\mathbf{x}_{1:T}^*):=\sum_{t=1}^T \|\mathbf{x}^*_t-\mathbf{x}^*_{t-1}\|$.

Since $\mathbb{E}[\|\bm{\lambda}_{T+1}\|^2]\geq 0$, initializing the dual variable with $\bm{\lambda}_1=\mathbf{0}$, and rearranging \eqref{eq.app1-12}, we have that
 \begin{align}\label{eq.app1-13}
  	&\sum_{t=1}^T\left(\mathbb{E}[\check{f}_t(\hat{\mathbf{x}}_t)]\!-\!\check{f}_t((1-\gamma)\mathbf{x}^*_t)\right)\nonumber\\
\!\!\leq &\gamma GR\|\bar{\bm{\lambda}}\|T\!+\!\frac{R}{\alpha}V(\mathbf{x}_{1:T}^*)\!+\!2\mu G^2R^2T\!+\!\frac{R^2}{2\alpha}\!+\!\frac{\alpha d^2F^2 T}{\delta^2}.
  \end{align}

 The iterates $\{\hat{\mathbf{x}}_t\}$ in this bound are not the actual decisions taken by the learner.
   To obtain the regret bound, our next step is to decompose the regret as
    \begin{align}\label{eq.app1-14}
    	\sum_{t=1}^T\Big(&\mathbb{E}[f_t(\mathbf{x}_{1,t})]-f_t(\mathbf{x}_t^*)\Big)
    =\sum_{t=1}^T\Big(\underbrace{\mathbb{E}[f_t(\mathbf{x}_{1,t})]-\mathbb{E}[\check{f}_t(\mathbf{x}_{1,t})]}_{U_1}\nonumber\\
    &+\underbrace{\mathbb{E}[\check{f}_t(\mathbf{x}_{1,t})]-\mathbb{E}[\check{f}_t(\hat{\mathbf{x}}_t)]}_{U_2}\!+\underbrace{\mathbb{E}[\check{f}_t(\hat{\mathbf{x}}_t)]-\check{f}_t((1-\gamma)\mathbf{x}_t^*))}_{U_3}\nonumber\\
    &+\underbrace{\check{f}_t((1-\gamma)\mathbf{x}_t^*))-\check{f}_t(\mathbf{x}_t^*)}_{U_4}+\underbrace{\check{f}_t(\mathbf{x}_t^*)-f_t(\mathbf{x}_t^*)}_{U_5}\Big).
    \end{align}
We next bound each under-braced, starting with
  \begin{align}\label{eq.app1-14a}
  	U_1=&\mathbb{E}\left[f_t(\mathbf{x}_{1,t})-\mathbb{E}_{\mathbf{v}}[f_t(\mathbf{x}_{1,t}+\delta\mathbf{v}_t)]\right]\nonumber\\
  	\stackrel{(d)}{\leq}& \mathbb{E}\left[f_t(\mathbf{x}_{1,t})-f_t(\mathbb{E}_{\mathbf{v}}[\mathbf{x}_{1,t}+\delta\mathbf{v}_t])\right]\stackrel{(e)}{=}0
 \end{align}
 where (d) uses Jensen's inequality, and (e) follows from $\mathbb{E}_{\mathbf{v}}[\delta\mathbf{v}_t]=\mathbf{0}$ since $\mathbf{v}_t$ is drawn from $\mathbb{B}:=\{\mathbf{v}:\|\mathbf{v}\|\leq 1\}$.
 
 Regarding the second term, it follows that   
 \begin{align}\label{eq.app1-14b}
    U_2=\mathbb{E}[\check{f}_t(\hat{\mathbf{x}}_t+\delta\mathbf{u}_t)-\check{f}_t(\hat{\mathbf{x}}_t)]
    	\stackrel{(f)}{\leq} \mathbb{E}[G\|\delta\mathbf{u}_t\|]=\delta G
\end{align}
where (f) uses the Lipschitz condition of $\check{f}_t(\mathbf{x})$.
The third term $U_3$ has been already bounded as in \eqref{eq.app1-13}. 

Using the Lipschitz condition of $\check{f}_t(\mathbf{x})$, we can further bound the fourth term
   \begin{align}\label{eq.app1-14d}
   	U_4=\check{f}_t((1-\gamma)\mathbf{x}_t^*))-\check{f}_t(\mathbf{x}_t^*)\leq \gamma GR
   \end{align}
   and likewise for the last term for which
 \begin{align}\label{eq.app1-14e}
 		U_5=\mathbb{E}_{\mathbf{v}}[f_t(\mathbf{x}_t^*+\delta\mathbf{v}_t)]\!-\!f_t(\mathbf{x}_t^*)
 		\leq \mathbb{E}_{\mathbf{v}}\left[G\|\delta\mathbf{v}_t\|\right]\!\leq\! \delta G.
 \end{align}
   
Plugging \eqref{eq.app1-13} and \eqref{eq.app1-14a}-\eqref{eq.app1-14e} into \eqref{eq.app1-14}, we arrive that
 \begin{align}\label{eq.app1-14all}
 \sum_{t=1}^T\Big(\mathbb{E}[f_t(\mathbf{x}_{1,t})]-f_t(\mathbf{x}_t^*)\Big) \leq \!\frac{R}{\alpha}V(\mathbf{x}_{1:T}^*)&+\!\frac{R^2}{2\alpha}+\!\frac{d^2G^2R^2\alpha T}{\delta^2}\nonumber\\
  +\gamma GRT(1+\|\bar{\bm{\lambda}}\|)+\!2\mu G^2R^2T&+\!2G\delta T.
 \end{align}
Upon choosing $\alpha=\mu={\cal O}(T^{-\frac{3}{4}})$, and $\delta={\cal O}(T^{-\frac{1}{4}})$ along with $\gamma=\delta/r$, it follows that (cf. Lemma \ref{app-lemma4})
 \begin{align*}
 	  {\rm Reg}^{\rm d}_T\!=\! {\cal O}\Big(\!RV(\mathbf{x}_{1:T}^*)T^{\frac{3}{4}}\!\!+\!GRCT^{\frac{3}{4}}\!\!+\!2G^2R^2T^{\frac{1}{4}}\!\!+\!d^2G^2R^2T^{\frac{3}{4}}\!\Big)
 \end{align*}
from which the proof is complete.

\textbf{Dynamic fit in Theorem 1:}
To bound the dynamic fit, recall that the constraint violations in \eqref{eq.prop1-1} depend on the magnitude of the dual variable and the difference of two consecutive primal iterates.
 The distance between iterates $\hat{\mathbf{x}}_t$ and $\hat{\mathbf{x}}_{t+1}$ can be bounded as
\begin{align}\label{eq.proof-fit0d}
&\|\hat{\mathbf{x}}_{t+1}-\hat{\mathbf{x}}_t\|\stackrel{(a)}{\leq} \big\|\alpha\hat{\nabla}_{\mathbf{x}}^1{\cal L}_t(\hat{\mathbf{x}}_t,\bm{\lambda}_t)\big\|\nonumber\\
\stackrel{(b)}{\leq} & \frac{d}{\delta}|f_t(\hat{\mathbf{x}}_t+\delta \mathbf{u}_t)|\!+\!\|\nabla \mathbf{g}_t(\hat{\mathbf{x}}_t)\|\|\bm{\lambda}_t\|\!\stackrel{(c)}{\leq}\! \frac{\alpha dF}{\delta}\!+\!\alpha G\|\bm{\lambda}_t\|
\end{align}
where (a) uses the non-expansive property of the projection operator, (b) relies on \eqref{eq.grad2} and the Cauchy-Schwarz's inequality; and (c) uses the bounds in (as2).

On the other hand, using the Lipschitz continuity of $\mathbf{g}_t(\mathbf{x})$ and \eqref{eq.prop1-1}, it follows that
\begin{align}\label{eq.proof-fit0}
	&\sum_{t=1}^T \mathbf{g}_t(\mathbf{x}_{1,t})\leq \sum_{t=1}^T \mathbf{g}_t(\hat{\mathbf{x}}_t)+\delta GT\mathbf{1}\\
	\leq &\frac{\bm{\lambda}_{T+1}}{\mu}+\frac{G^2T\mathbf{1}}{2\beta}+\frac{\beta}{2}\sum_{t=1}^T\|\hat{\mathbf{x}}_{t+1}-\hat{\mathbf{x}}_t\|^2\mathbf{1}+\delta GT\mathbf{1}\nonumber\\
	\stackrel{(d)}{\leq}&\frac{\bm{\lambda}_{T+1}}{\mu}+\frac{G^2T\mathbf{1}}{2\beta}+\beta T\Big(\frac{\alpha^2 d^2F^2}{\delta^2}+\alpha^2 G^2\|\bar{\bm{\lambda}}\|^2\Big)\mathbf{1}+\delta GT\mathbf{1}\nonumber
\end{align}
where (d) uses \eqref{eq.proof-fit0d}, and the fact that $(a+b)^2\leq 2(a^2+b^2)$. 
Taking $[\cdot]^+$ and $\|\cdot\|$ on both sides of \eqref{eq.proof-fit0}, we have (cf. \eqref{eq.dyn-fit})
\begin{align}\label{eq.app-Them.fit}
        \!{\rm Fit}^{\rm d}_T\!\leq\! \frac{\|\bar{\bm{\lambda}}\|}{\mu}&+\!\frac{G^2\sqrt{N}T}{2\beta}\!+\!\delta G\sqrt{N}T\nonumber\\
        &+\!\beta \sqrt{N} T\left(\!\frac{\alpha^2 d^2F^2 }{\delta^2}\!+\!\alpha^2 G^2\|\bar{\bm{\lambda}}\|^2\!\right)\!\!
\end{align}
which establishes \eqref{Them.fit}.
Upon selecting $\alpha={\cal O}(T^{-\frac{3}{4}})$, and $\delta={\cal O}(T^{-\frac{1}{4}})$, we find from Lemma \ref{app-lemma4} that $\|\bar{\bm{\lambda}}\|\leq C={\cal O}(1)$. Together with $\mu={\cal O}(T^{-\frac{3}{4}})$ and $\beta={\cal O}(T^{\frac{1}{4}})$, it holds from \eqref{eq.app-Them.fit} that
\begin{align}\label{eq.proof-fit}
        \!{\rm Fit}^{\rm d}_T\leq & CT^{\frac{3}{4}}+\!\frac{G^2\sqrt{N}T^{\frac{3}{4}}}{2}\!+\!G\sqrt{N}T^{\frac{3}{4}}\nonumber\\
        &+\!\sqrt{N} T^{\frac{5}{4}}\left(d^2F^2 T^{-1}\!+\!T^{-\frac{3}{2}} G^2C^2\!\right)={\cal O}(T^{\frac{3}{4}})
\end{align}
which completes the proof of \eqref{Them.dyn-reg1}.

\subsection{Proof of Theorem \ref{Them2}}\label{app-prf2}
Similar to the proof of Theorem \ref{Them1}, feasibility of actions $\{\mathbf{x}_{1,t},\mathbf{x}_{2,t}\}$ readily follows from Lemma \ref{app-lemma1}; hence, $\mathbf{x}_{1,t},\mathbf{x}_{2,t}\in{\cal X},\,\forall t$. 
To prove the dynamic regret and fit bounds in this setup, the following result is needed.
 \begin{lemma}\label{app-lemma5}
	For the BanSaP recursion \eqref{eq.primal2}, \eqref{eq.dual2}, and \eqref{eq.grad2}, selecting $\alpha=\mu={\cal O}(T^{-\frac{1}{2}})$ ensures that the dual iterates are uniformly bounded by $\|\bm{\lambda}_t\|\leq C={\cal O}(1)$, with the constant $C$ given by
\begin{equation}\label{eq.defi-C2}
	\!C\!:=\!\max\left\{\!2GR,\left(\frac{1}{\eta}+1\right)\!GR\!+\!\frac{2G^2R^2\mu}{\eta}\!+\!\frac{d^2G^2\alpha}{\eta}\!+\!\frac{\mu R^2}{2\alpha\eta}\!\right\}\!
\end{equation}
where the constants $G$, $R$, and $\eta$ are as in (as2)-(as4).
\end{lemma}

\begin{IEEEproof}
It follows steps similar to those used to prove Lemma \ref{app-lemma4}.
\end{IEEEproof}

Similar to Lemma \ref{app-lemma4}, Lemma \ref{app-lemma5} asserts that the dual variable in BanSaP with two-point bandit feedback is also uniformly bounded from above. 
Now, we are ready to prove the regret bound in Theorem \ref{Them2}.

\textbf{Dynamic regret in Theorem 2:}
   To obtain the regret bound in the case of two-point feedback, our first step is to connect the regret with the optimality loss induced by the sequence of iterates $\{\hat{\mathbf{x}}_t\}$, given by 
       \begin{align}\label{eq.app2-14}
       	    &	\frac{1}{2}\sum_{t=1}^T\Big(\mathbb{E}[f_t(\mathbf{x}_{1,t})]+\mathbb{E}[f_t(\mathbf{x}_{2,t})]\Big)-\sum_{t=1}^Tf_t(\mathbf{x}_t^*)\nonumber\\
    \stackrel{(a)}{\leq}&\frac{1}{2}\sum_{t=1}^T\Big(\mathbb{E}[f_t(\hat{\mathbf{x}}_t)]+\delta G+\mathbb{E}[f_t(\hat{\mathbf{x}}_t)]+\delta G\Big)-\sum_{t=1}^Tf_t(\mathbf{x}_t^*)\nonumber\\
    =&\sum_{t=1}^T\Big(\mathbb{E}[f_t(\hat{\mathbf{x}}_t)]-f_t(\mathbf{x}_t^*)\Big)+\delta G
     \end{align}
     where (a) follows from the Lipschitz condition in (as2).

The LHS of \eqref{eq.app2-14} can be further decomposed as
    \begin{align}\label{eq.app2-14b}
    	&\sum_{t=1}^T\Big(\mathbb{E}[f_t(\hat{\mathbf{x}}_t)]-f_t(\mathbf{x}_t^*)\Big)\nonumber\\
    =&\sum_{t=1}^T\Big(\underbrace{\mathbb{E}[f_t(\hat{\mathbf{x}}_t)]-\mathbb{E}[\check{f}_t(\hat{\mathbf{x}}_t)]}_{U_1}+\underbrace{\mathbb{E}[\check{f}_t(\hat{\mathbf{x}}_t)]-\check{f}_t((1-\gamma)\mathbf{x}_t^*))}_{U_2}\nonumber\\
    &\qquad~+\underbrace{\check{f}_t((1-\gamma)\mathbf{x}_t^*))\!-\!\check{f}_t(\mathbf{x}_t^*)}_{U_3}+\underbrace{\check{f}_t(\mathbf{x}_t^*)\!-\!f_t(\mathbf{x}_t^*)}_{U_4}\Big).
    \end{align}
    
For the first term, following the steps in \eqref{eq.app1-14a}, we have that
  \begin{align}\label{eq.app2-15a}
  	U_1{\leq}&\mathbb{E}\left[f_t(\hat{\mathbf{x}}_t)-\mathbb{E}_{\mathbf{v}}[f_t(\hat{\mathbf{x}}_t+\delta\mathbf{v}_t)]\right]\nonumber\\
  	{\leq}& \mathbb{E}\left[f_t(\hat{\mathbf{x}}_t)-f_t(\mathbb{E}_{\mathbf{v}}[\hat{\mathbf{x}}+\delta\mathbf{v}_t])\right]\leq0.
 \end{align}
   
Similar to \eqref{eq.app1-13}, we have for the case of two-point feedback
 \begin{align}\label{eq.app2-15b}
 	\sum_{t=1}^TU_2&=\sum_{t=1}^T\left(\mathbb{E}[\check{f}_t(\hat{\mathbf{x}}_t)]\!-\!\check{f}_t((1-\gamma)\mathbf{x}^*_t)\right)\\
&\!\leq\! \gamma GR\|\bar{\bm{\lambda}}\|T\!+\!\frac{R}{\alpha}V(\mathbf{x}_{1:T}^*)\!+\!2\mu G^2R^2T\!+\!\frac{R^2}{2\alpha}\!+\!\alpha d^2G^2 T.\nonumber
  \end{align}
  
Using the Lipschitz condition of $\check{f}_t(\mathbf{x})$, we can bound the third term
   \begin{align}\label{eq.app2-15c}
   	U_3=\check{f}_t((1-\gamma)\mathbf{x}_t^*))-\check{f}_t(\mathbf{x}_t^*)\leq \gamma GR
   \end{align}
   and likewise for the last term, it follows from the Lipschitz condition of $f_t(\mathbf{x})$ that
 \begin{align}\label{eq.app2-15d}
 		U_4=\mathbb{E}_{\mathbf{v}}[f_t(\mathbf{x}_t^*+\delta\mathbf{v}_t)]\!-\!f_t(\mathbf{x}_t^*)\leq \delta G.
 \end{align}

Plugging \eqref{eq.app2-15a}-\eqref{eq.app2-15d} into \eqref{eq.app2-14}, we arrive at
       \begin{align}\label{eq.app2-16}
    \!\!\!\! 	    &	\frac{1}{2}\sum_{t=1}^T\Big(\mathbb{E}[f_t(\mathbf{x}_{1,t})]+\mathbb{E}[f_t(\mathbf{x}_{2,t})]\Big)-\sum_{t=1}^Tf_t(\mathbf{x}_t^*) \leq \!\frac{R}{\alpha}V(\mathbf{x}_{1:T}^*)\!\nonumber\\
  \!\!\!\!\!\!&\!+\!\frac{R^2}{2\alpha}\!+\!2\mu G^2R^2T\!+\!\alpha d^2G^2 T\!+\!\gamma GRT(1\!+\!\|\bar{\bm{\lambda}}\|)\!+\!2\delta GT. \!\!
     \end{align}
     
Upon choosing $\alpha=\mu={\cal O}(T^{-\frac{1}{2}})$, and $\delta={\cal O}(T^{-1})$ along with $\gamma=\delta/r$, it follows that (ignoring constant terms)
 \begin{align*}
 	  {\rm Reg}^{\rm d}_T\!=\! {\cal O}\Big(RV(\mathbf{x}_{1:T}^*)T^{\frac{1}{2}}\!+\frac{1}{2}R^2T^{\frac{1}{2}}\!+\!2G^2R^2T^{\frac{1}{2}}\!\!+\!d^2G^2T^{\frac{1}{2}}\Big)
 \end{align*}
where we used the upper bound of dual variables in Lemma \ref{app-lemma5}. This completes the proof of \eqref{Them2.dyn-reg}.

\textbf{Dynamic fit in Theorem 2:}
To derive the bound on dynamic fit, recall that the constraint violations in \eqref{eq.prop1-1} depend on the magnitude of the dual variable as well as on the difference of two consecutive primal iterates.
 The distance between iterates $\mathbf{x}_t$ and $\hat{\mathbf{x}}_{t+1}$ can be bounded by
\begin{align}\label{eq.proof-fit1d}
&\|\hat{\mathbf{x}}_{t+1}-\hat{\mathbf{x}}_t\|\stackrel{(a)}{\leq} \big\|\alpha\hat{\nabla}_{\mathbf{x}}^2{\cal L}_t(\hat{\mathbf{x}}_t,\bm{\lambda}_t)\big\|\nonumber\\
\stackrel{(b)}{\leq} & \|\hat{\nabla}^2 f_t(\hat{\mathbf{x}}_t)\|+\|\nabla \mathbf{g}_t(\hat{\mathbf{x}}_t)\|\|\bm{\lambda}_t\|\stackrel{(c)}{\leq} \alpha dG+\alpha G \|\bar{\bm{\lambda}}\|
\end{align}
where (a) uses the non-expansive property of the projection operator, (b) applies the Cauchy-Schwarz inequality, and (c) relies on the bounds in (as2).

On the other hand, using the Lipschitz continuity of $\mathbf{g}_t(\mathbf{x})$ and \eqref{eq.prop1-1}, we have
\begin{align}\label{eq.proof-fit1}
	&\frac{1}{2}\sum_{t=1}^T \left(\mathbf{g}_t(\mathbf{x}_{1,t})+\mathbf{g}_t(\mathbf{x}_{2,t})\right)\leq \sum_{t=1}^T \mathbf{g}_t(\hat{\mathbf{x}}_t)+\delta GT\mathbf{1}\\
	\leq &\frac{\bm{\lambda}_{T+1}}{\mu}+\frac{G^2T\mathbf{1}}{2\beta}+\frac{\beta}{2}\sum_{t=1}^T\|\hat{\mathbf{x}}_{t+1}-\hat{\mathbf{x}}_t\|^2\mathbf{1}+\delta GT\mathbf{1}\nonumber\\
	\stackrel{(c)}{\leq}&\frac{\bm{\lambda}_{T+1}}{\mu}+\frac{G^2T\mathbf{1}}{2\beta}+\beta T\Big(\alpha^2 d^2G^2+\alpha^2 G^2\|\bar{\bm{\lambda}}\|^2\Big)\mathbf{1}+\delta GT\mathbf{1}\nonumber
\end{align}
where (c) uses \eqref{eq.proof-fit1d}, and the fact that $(a+b)^2\leq 2(a^2+b^2)$. 

In this case, if we take $[\cdot]^+$ and then $\|\cdot\|$ on both sides of \eqref{eq.proof-fit1}, and further choose $\alpha=\mu={\cal O}(T^{-\frac{1}{2}})$, $\delta=T^{-1}$, and $\beta={\cal O}(T^{\frac{1}{2}})$, we arrive at
\begin{align}\label{eq.proof-fit2}
{\rm Fit}^{\rm d}_T\leq&\frac{\|\bm{\lambda}_{T+1}\|}{\mu}+\frac{G^2N^{\frac{1}{2}}T}{2\beta}\!+\!\beta N^{\frac{1}{2}}T\Big(\alpha^2 d^2G^2+\alpha^2 G^2\|\bar{\bm{\lambda}}\|^2\Big)\nonumber\\
=&CT^{\frac{1}{2}}+N^{\frac{1}{2}}T^{\frac{1}{2}}G^2\left(\frac{1}{2}+d^2+C^2\right)={\cal O}\left(T^{\frac{1}{2}}\right)
\end{align}
where we used the bound on dual variables in Lemma \ref{app-lemma5}. This completes also the proof of \eqref{Them2.dyn-reg1}, and also that of Theorem \ref{Them2}.



\begin{thebibliography}{10}
\providecommand{\url}[1]{#1}
\csname url@samestyle\endcsname
\providecommand{\newblock}{\relax}
\providecommand{\bibinfo}[2]{#2}
\providecommand{\BIBentrySTDinterwordspacing}{\spaceskip=0pt\relax}
\providecommand{\BIBentryALTinterwordstretchfactor}{4}
\providecommand{\BIBentryALTinterwordspacing}{\spaceskip=\fontdimen2\font plus
\BIBentryALTinterwordstretchfactor\fontdimen3\font minus
  \fontdimen4\font\relax}
\providecommand{\BIBforeignlanguage}[2]{{%
\expandafter\ifx\csname l@#1\endcsname\relax
\typeout{** WARNING: IEEEtran.bst: No hyphenation pattern has been}%
\typeout{** loaded for the language `#1'. Using the pattern for}%
\typeout{** the default language instead.}%
\else
\language=\csname l@#1\endcsname
\fi
#2}}
\providecommand{\BIBdecl}{\relax}
\BIBdecl

\bibitem{samie2016b}
F.~Samie, V.~Tsoutsouras, S.~Xydis, L.~Bauer, D.~Soudris, and J.~Henkel,
  ``Distributed {QoS} management for {Internet of Things} under resource
  constraints,'' in \emph{Proc. Intl. Conf. on Hardware/Software Codesign and
  System Synthesis}, Pittsburgh, PA, Oct. 2016, pp. 1--10.

\bibitem{chiang2016}
M.~Chiang and T.~Zhang, ``Fog and {IoT}: An overview of research
  opportunities,'' \emph{IEEE Internet Things J.}, vol.~3, no.~6, pp. 854--864,
  2016.

\bibitem{lee2017}
G.~Lee, W.~Saad, and M.~Bennis, ``An online secretary framework for fog network
  formation with minimal latency,'' \emph{arXiv:1702.05569}, Apr. 2017.

\bibitem{samie2016}
F.~Samie, V.~Tsoutsouras, L.~Bauer, S.~Xydis, D.~Soudris, and J.~Henkel,
  ``Computation offloading and resource allocation for low-power {IoT} edge
  devices,'' in \emph{Proc. World Forum Internet Things}, Dec. 2016, pp. 7--12.

\bibitem{mach2017}
P.~Mach and Z.~Becvar, ``Mobile edge computing: {A} survey on architecture and
  computation offloading,'' \emph{IEEE Comm. Surveys \& Tutorials}, 2017, to
  appear.

\bibitem{wang2017}
\BIBentryALTinterwordspacing
F.~Wang, J.~Xu, X.~Wang, and S.~Cui, ``Joint offloading and computing
  optimization in wireless powered mobile-edge computing systems,''
  \emph{{IEEE} Trans. Wireless Commun.}, Feb. 2017, submitted. [Online].
  Available: \url{https://arxiv.org/abs/1702.00606}
\BIBentrySTDinterwordspacing

\bibitem{mao2017}
Y.~Mao, C.~You, J.~Zhang, K.~Huang, and K.~B. Letaief, ``Mobile edge computing:
  {S}urvey and research outlook,'' \emph{arXiv preprint:1701.01090}, Jan. 2017.

\bibitem{zinkevich2003}
M.~Zinkevich, ``Online convex programming and generalized infinitesimal
  gradient ascent,'' in \emph{Proc. Intl. Conf. on Machine Learning},
  Washington D.C., Aug. 2003.

\bibitem{hazan2007}
E.~Hazan, A.~Agarwal, and S.~Kale, ``Logarithmic regret algorithms for online
  convex optimization,'' \emph{Machine Learning}, vol.~69, no. 2-3, pp.
  169--192, Dec. 2007.

\bibitem{duchi2011jmlr}
J.~C. Duchi, E.~Hazan, and Y.~Singer, ``Adaptive subgradient methods for online
  learning and stochastic optimization,'' \emph{Journal of Machine Learning
  Research}, vol.~12, pp. 2121--2159, Jul. 2011.

\bibitem{jadbabaie2015}
A.~Jadbabaie, A.~Rakhlin, S.~Shahrampour, and K.~Sridharan, ``Online
  optimization: Competing with dynamic comparators,'' in \emph{Intl. Conf. on
  Artificial Intelligence and Statistics}, San Diego, CA, May 2015.

\bibitem{hall2015}
E.~C. Hall and R.~M. Willett, ``Online convex optimization in dynamic
  environments,'' \emph{{IEEE} J. Sel. Topics Signal Process.}, vol.~9, no.~4,
  pp. 647--662, Jun. 2015.

\bibitem{chen2017tsp}
T.~Chen, Q.~Ling, and G.~B. Giannakis, ``An online convex optimization approach
  to proactive network resource allocation,'' \emph{{IEEE} Trans. Signal
  Processing}, Jan. 2017 (revised), {Available:
  https://arxiv.org/abs/1701.03974}.

\bibitem{chen2017iot}
\BIBentryALTinterwordspacing
T.~Chen, Y.~Shen, Q.~Ling, and G.~B. Giannakis, ``Online learning for
  ``thing-adaptive'' fog computing in {IoT},'' in \emph{Proc. of Asilomar
  Conf.}, Pacific Grove, CA, Oct. 2017. [Online]. Available:
  \url{www.dropbox.com/s/z4qnog6x0gzd2ko/TAOSP.pdf?dl=0}
\BIBentrySTDinterwordspacing

\bibitem{neely2017}
M.~J. Neely and H.~Yu, ``Online convex optimization with time-varying
  constraints,'' \emph{arXiv preprint:1702.04783}, Feb. 2017.

\bibitem{flaxman2005}
A.~D. Flaxman, A.~T. Kalai, and H.~B. McMahan, ``Online convex optimization in
  the bandit setting: gradient descent without a gradient,'' in \emph{Proc. of
  ACM SODA}, Vancouver, Canada, Jan. 2005, pp. 385--394.

\bibitem{agarwal2010}
A.~Agarwal, O.~Dekel, and L.~Xiao, ``Optimal algorithms for online convex
  optimization with multi-point bandit feedback.'' in \emph{Proc. Annual Conf.
  on Learning Theory}, Haifa, Israel, 2010, pp. 28--40.

\bibitem{shamir2017}
O.~Shamir, ``An optimal algorithm for bandit and zero-order convex optimization
  with two-point feedback,'' \emph{Journal of Machine Learning Research},
  vol.~18, no.~52, pp. 1--11, 2017.

\bibitem{bubeck2012}
S.~Bubeck and N.~Cesa-Bianchi, ``Regret analysis of stochastic and
  nonstochastic multi-armed bandit problems,'' \emph{Found. and Trends in Mach.
  Learn.}, vol.~5, no.~1, pp. 1--122, 2012.

\bibitem{duchi2015}
J.~C. Duchi, M.~I. Jordan, M.~J. Wainwright, and A.~Wibisono, ``Optimal rates
  for zero-order convex optimization: The power of two function evaluations,''
  \emph{{IEEE} Trans. Inform. Theory}, vol.~61, no.~5, pp. 2788--2806, May
  2015.

\bibitem{nesterov2017}
Y.~Nesterov and V.~Spokoiny, ``Random gradient-free minimization of convex
  functions,'' \emph{Foundations of Computational Mathematics}, vol.~17, no.~2,
  pp. 527--566, Apr. 2017.

\bibitem{mahdavi2012}
M.~Mahdavi, R.~Jin, and T.~Yang, ``Trading regret for efficiency: {O}nline
  convex optimization with long term constraints,'' \emph{Journal of Machine
  Learning Research}, vol.~13, pp. 2503--2528, Sep 2012.

\bibitem{awerbuch2004}
B.~Awerbuch and R.~D. Kleinberg, ``Adaptive routing with end-to-end feedback:
  Distributed learning and geometric approaches,'' in \emph{Proc. ACM Symp. on
  Theory of Computing}, Chicago, IL, Jun. 2004, pp. 45--53.

\bibitem{kao2016}
\BIBentryALTinterwordspacing
Y.-H. Kao, K.~Wright, B.~Krishnamachari, and F.~Bai, ``Online learning for
  wireless distributed computing,'' \emph{arXiv preprint:1611.02830}, Nov.
  2016. [Online]. Available: \url{https://arxiv.org/abs/1611.02830}
\BIBentrySTDinterwordspacing

\bibitem{neely2010}
M.~J. Neely, ``Stochastic network optimization with application to
  communication and queueing systems,'' \emph{Synthesis Lectures on
  Communication Networks}, vol.~3, no.~1, pp. 1--211, 2010.

\bibitem{chen2016tsp}
T.~Chen, A.~Mokhtari, X.~Wang, A.~Ribeiro, and G.~B. Giannakis, ``Stochastic
  averaging for constrained optimization with application to online resource
  allocation,'' \emph{{IEEE} Trans. Signal Processing}, vol.~65, no.~12, pp.
  3078--3093, Jun. 2017.

\bibitem{sardellitti2015}
S.~Sardellitti, G.~Scutari, and S.~Barbarossa, ``Joint optimization of radio
  and computational resources for multicell mobile-edge computing,'' \emph{IEEE
  Trans. Signal Info. Process. Netw.}, vol.~1, no.~2, pp. 89--103, Jun. 2015.

\bibitem{chen2016ton}
X.~Chen, L.~Jiao, W.~Li, and X.~Fu, ``Efficient multi-user computation
  offloading for mobile-edge cloud computing,'' \emph{{IEEE/ACM} Trans.
  Networking}, vol.~24, no.~5, pp. 2795--2808, Oct. 2016.

\bibitem{huang2017}
H.~Huang, Q.~Ling, W.~Shi, and J.~Wang, ``Collaborative resource allocation
  over a hybrid cloud center and edge server network,'' \emph{Journal of
  Computational Mathematics}, 2017, to appear.

\bibitem{besbes2015}
O.~Besbes, Y.~Gur, and A.~Zeevi, ``Non-stationary stochastic optimization,''
  \emph{Operations Research}, vol.~63, no.~5, pp. 1227--1244, Sep. 2015.

\bibitem{bertsekas1999}
D.~P. Bertsekas, \emph{Nonlinear Programming}.\hskip 1em plus 0.5em minus
  0.4em\relax Belmont, MA: Athena scientific, 1999.

\end{thebibliography}
\end{document}